\documentclass[11pt,twoside]{article}
\usepackage[affil-it]{authblk}
\usepackage[T1]{fontenc}
\usepackage{amsmath}
\usepackage{amsfonts}
\usepackage{amssymb}
\usepackage{graphicx}
\usepackage{float}
\usepackage{booktabs}
\usepackage{csquotes}
\usepackage{array}
\usepackage{multicol}
\usepackage{fancyhdr}
\usepackage{geometry}
\usepackage[hang,ragged]{footmisc}
\usepackage{hyperref}
\usepackage{url}
\usepackage[style=authoryear,maxbibnames=10]{biblatex}
\usepackage{tabulary}
\usepackage{tabularx}
\usepackage{caption}
\usepackage{titlesec}
\usepackage[protrusion=true,expansion=false,tracking=true,nopatch=footnote]{microtype}
\usepackage[dvipsnames]{xcolor}
\usepackage{colortbl}
\usepackage[oldstyle,proportional]{libertinus}
\usepackage[ttdefault]{sourcecodepro}
\usepackage{FiraSans}

\newlength{\gridunit}
\normalsize
\newcount\gridlines
\gridlines=\dimexpr\textheight-\topskip+0.5\gridunit\relax
\divide\gridlines by \gridunit
\titleformat*{\section}{\sffamily\bfseries\large\raggedright\lineskiplimit=-0.5\gridunit\setlength{\baselineskip}{\gridunit}}
\titleformat*{\subsection}{\sffamily\bfseries\normalsize\raggedright\lineskiplimit=-0.5\gridunit\setlength{\baselineskip}{\gridunit}}
\titleformat*{\subsubsection}{\sffamily\bfseries\small\raggedright\lineskiplimit=-0.5\gridunit\setlength{\baselineskip}{\gridunit}}
\titleformat*{\paragraph}{\sffamily\bfseries\footnotesize\raggedright}

\titlespacing*{\section}{0pt}{\gridunit}{\gridunit}
\titlespacing*{\subsection}{0pt}{\gridunit}{\gridunit}
\titlespacing*{\subsubsection}{0pt}{\gridunit}{\gridunit}
\titlespacing*{\paragraph}{0pt}{\gridunit}{0.5em}

\DeclareTextFontCommand{\token}{\footnotesize\sffamily}
\DeclareTextFontCommand{\tokenfn}{\scriptsize\sffamily}
\titlelabel{\thetitle.\enspace}

\fancypagestyle{firststyle}{
  \fancyhf{}
  \fancyfoot[C]{\sffamily\scriptsize \thepage}
}

\DeclareCaptionFormat{base}{\scriptsize\sffamily\ \textbf{#1#2}{#3}}
\DefineBibliographyStrings{english}{page={},pages={}}

\DeclareFieldFormat{postnote}{#1}
\DeclareBibliographyCategory{needsurl}
\renewbibmacro*{url+urldate}{%
  \ifcategory{needsurl}
  {\printfield{url}%
    \iffieldundef{urlyear}
    {}
    {\setunit*{\addspace}%
  \printurldate}}
{}}
\newcommand{\entryneedsurl}[1]{\addtocategory{needsurl}{#1}}
\entryneedsurl{smailDALMEDocumentaryArchaeology2020}
\entryneedsurl{stempelDictionnaireLOccitanMedieval2012}
\DeclareCiteCommand{\citerecord}[\mkbibfootnote]{}
{\printfield[citetitle]{labeltitle}
  \addspace\mkbibparens{\printfield{postnote}}
  \multinamedelim\addspace\usebibmacro{cite}
\addcolon\space\printfield[citeurl]{url}}
{\multicitedelim}{}
\DeclareMultiCiteCommand{\citerecords}[\mkbibfootnote]{\citerecord}{\multicitedelim}

\hypersetup{
  colorlinks=true,
  linkcolor=black,
  filecolor=black,
  urlcolor=black,
  citecolor=black,
}
\title{Correction as Annotation: \\ Bootstrapping a Dependency Parser \\ for Documentary Medieval Latin\vspace{20pt}}
\author{Gabriel H. Pizzorno\textsuperscript{*}}
\affil{Harvard University\vspace{-20pt}}
\date{}

\begin{document}

\maketitle
\thispagestyle{firststyle}

\begin{abstract}
  \itshape
  Medieval documentary sources remain inadequately served by existing natural language processing tools. None of the five readily available Latin treebank models attains usable performance on a collection of 160 inventories compiled in Marseille between 1258 and 1446. The best labelled attachment score is 0.62 and the best morphology-aware score is 0.24. Performance does not correlate with either genre or period proximity. To address this shortfall, in-domain training data was generated as a by-product of using these inadequate models. In each of nine iterations, a model pre-annotated 200 sentences; an expert corrected the annotations; and the corrected sentences were used to train the subsequent model, with batches sampled independently of model state, without active-learning selection. Thirty-three hours of annotation effort over 1,804 sentences increased universal part-of-speech accuracy from 0.80 to 0.98 and labelled attachment from 0.48 to 0.92, outperforming all baselines on the reported metrics while using 97\% less training data than the largest one of them. Annotator effort declined from 54\% of tokens to a plateau of 14–18\%, an operational progress metric that requires no separate gold standard and can serve as a stopping criterion.
  \vspace{10pt}
\end{abstract}

\vspace{10pt}

\begin{multicols}{2}

  \renewcommand{\thefootnote}{*}
  \footnotetext{Department of History | Robinson Hall, Office 210 | 35 Quincy Street, Cambridge, MA 02138 | \href{mailto:pizzorno@fas.harvard.edu}{pizzorno@fas.harvard.edu}}
  \renewcommand{\thefootnote}{\arabic{footnote}}

  \section{Introduction}

  The application of natural language processing (NLP) methods to historical sources presents unique challenges not typically encountered with contemporary texts. While the field has achieved remarkable success on languages that possess extensive corpora and standardized orthographies, historical texts remain difficult to process because of their linguistic idiosyncrasies, the limited availability of suitable training data, and their frequent divergence from both classical and modern variants of the same language \autocite{piotrowskiNaturalLanguageProcessing2012,kestemontLemmatizationVariationrichLanguages2017}.

  The challenges of historical NLP are especially acute for highly specialized corpora that exhibit domain-specific vocabularies, non-standard orthographies, and/or syntactic patterns that diverge markedly from the literary genres typically represented in existing treebanks. Such corpora frequently arise in documentary projects, where texts reflect specialized administrative practices and local linguistic variation that are insufficiently represented in standard resources \autocite{craneWhatYouMillion2006,mcgillivrayDigitalHumanitiesNatural2020}.

  Medieval documentary sources are an emblematic example, since they often combine Latin and vernacular elements, employ specialized technical vocabulary, and adhere to syntactic conventions shaped by administrative and legal discourse \autocite[350-351]{wrightSociophilologicalStudyLate2002}. For such corpora, conventional historical-NLP approaches—including the application of models pre-trained on existing treebanks or straightforward domain-adaptation techniques—frequently produce inadequate results \autocite{bammanAncientGreekLatin2011, passarottiLemlat30Package2017}.

  To address these challenges, this paper introduces a human-in-the-loop iterative training methodology designed to produce accurate NLP models for highly idiosyncratic historical corpora while minimizing manual annotation effort. The method pairs automated preprocessing by existing models with targeted human correction in an iterative bootstrapping loop, progressively improving model performance while maintaining strict quality control over annotations.

  The methodology is illustrated with a case study based on a collection of late-medieval inventories from Marseille (1258–1446). These inventories, produced in legal and administrative contexts, contain a mixture of Medieval Latin and vernacular Occitan\footnote{For conciseness, \emph{Occitan}, rather than \emph{Provençal} or \emph{Old Occitan}, is used throughout this paper.} terms, employ formulaic descriptive structures, and are linguistically distant from the varieties represented in existing Latin treebanks.

  The evaluation in Section~\ref{sec:baselines} shows that models trained on standard Latin treebanks attain only 47–57\% accuracy on the target corpus when a token's complete morphological analysis must be correct, and 15–24\% when attachment, label, and morphology must all be correct simultaneously.\footnote{F1 for All Tags (which requires the universal part-of-speech tag, the language-specific part-of-speech tag, the morphological features, and the lemma to be simultaneously correct) and for MLAS.} The human-in-the-loop methodology presented in this work allows for substantial improvements in model performance while requiring significantly less manual annotation than traditional treebank construction approaches.

  This paper makes three principal contributions. First, it provides a systematic analysis of the linguistic challenges posed by specialized historical corpora and demonstrate the inadequacy of existing models for such texts. Second, it introduces a human-in-the-loop iterative training methodology that efficiently adapts NLP models to corpora exhibiting extreme domain distance. Third, it presents a detailed case study applying this methodology to late-medieval documentary sources and offer practical guidance for analogous projects working with challenging historical corpora.

  \section{Related Work}

  The work in this paper intersects with several research areas in computational linguistics, notably domain adaptation, human-in-the-loop learning, and the computational processing of historical languages. The following situates the present contribution within those literatures.

  \subsection{Domain Adaptation for NLP}

  Domain adaptation remains a central problem in NLP as models trained on one domain typically degrade when applied to another \autocite{daumeiiiFrustratinglyEasyDomain2007,ben-davidTheoryLearningDifferent2010}. The problem is particularly pronounced for historical text, where differences in vocabulary, syntax, and orthographic conventions between source and target data can be substantial \autocite{raysonTaggingBardEvaluating2007,baronVARD2Tool2008}.

  Standard adaptation strategies include feature-based methods that aim to learn domain-invariant representations \autocite{blitzerDomainAdaptationStructural2006}, fine-tuning of pre-trained models on target data \autocite{howardUniversalLanguageModel2018}, and transfer-learning approaches that leverage related source domains \autocite{panSurveyTransferLearning2010}. These techniques typically presuppose a degree of similarity between source and target domains and therefore struggle when that assumption fails, as is often the case for specialized historical corpora.

  More recent work has explored techniques better suited to large domain gaps, including adversarial domain adaptation \autocite{ganinUnsupervisedDomainAdaptation2015} and multitask frameworks that jointly optimize for multiple related tasks \autocite{caruanaMultitaskLearning1997}. Character-level neural models have proved effective for historical-text normalization \autocite{bollmannLargescaleComparisonHistorical2019}, while careful preprocessing and adaptation of transformer-based encoders can yield further gains for historical language varieties \autocite{manjavacasAdaptingVsPretraining2022}.

  \subsection{Human-in-the-Loop Learning}

  Human-in-the-loop learning is a paradigm in which human expertise is systematically integrated into the machine learning process in order to improve model performance and reliability \autocite{holzingerInteractiveMachineLearning2016}. This approach has been shown to be particularly valuable when working with specialized domains where automated methods alone are often insufficient to achieve satisfactory results \autocite{amershiPowerPeopleRole2014}.

  Active learning, the best-developed component of this literature, selects informative examples for annotation according to criteria such as uncertainty sampling \autocite{lewisSequentialAlgorithmTraining1994}, query-by-committee \autocite{seungQueryCommittee1992}, or information-theoretic objectives \autocite{mackayInformationBasedObjectiveFunctions1992}. Extensions of these strategies to neural models have been explored extensively \autocite{settlesActiveLearningLiterature2009,galDropoutBayesianApproximation2015,shenDeepActiveLearning2017}. In NLP, human-in-the-loop methods have been applied to tasks including named-entity recognition, parsing, and machine translation \autocite{tomanekSemisupervisedActiveLearning2009,hwaBootstrappingParsersSyntactic2005,greenEfficacyHumanPostediting2013}. analyzes of cost/quality trade-offs and annotation noise have highlighted practical constraints in real-world annotation projects \autocite{fortAmazonMechanicalTurk2011,rehbeinDetectingAnnotationNoise2017}.

  \subsection{Historical Language Processing}

  Computational processing of historical languages has matured into a substantive research area concerned with orthographic variation, morphological complexity, and data scarcity \autocite{piotrowskiNaturalLanguageProcessing2012,kestemontLemmatizationVariationrichLanguages2017}.

  Early work in this area focused primarily on text normalization and spelling standardization \autocite{baronVARD2Tool2008,jurishMoreWordsUsing2010}, while more recent approaches have tackled complex NLP tasks such as parsing and semantic analysis. The development of historical treebanks has been crucial for advancing this field, including projects such as the Penn-Helsinki Parsed Corpora of Historical English \autocite{krochPennHelsinkiParsedCorpora2010}, the PROIEL treebank for ancient Indo-European languages \autocite{haugCreatingParallelTreebank2008}, and various Latin treebanks \autocite{bammanAncientGreekLatin2011,passarottiProjectIndexThomisticus2019}.

  Specific challenges in historical Latin processing have been addressed by several research efforts. The Index Thomisticus Treebank project \autocite{passarottiProjectIndexThomisticus2019,passarottiImprovementsParsingIndex2010} demonstrated the feasibility of parsing medieval Latin texts, while the Late Latin Charter Treebank \autocite{korkiakangasLateLatinCharter2021} explored the linguistic characteristics of early medieval documentary sources. These projects have revealed substantial variation in Latin across different time periods, registers, and geographical regions, highlighting the need for specialized approaches to different historical corpora.

  \subsection{Treebank Creation and Adaptation}

  The creation and adaptation of syntactic treebanks represents a fundamental challenge in NLP, requiring considerable expertise and resources \autocite{marcusBuildingLargeAnnotated1993,abeilleTreebanksBuildingUsing2003}. Traditional approaches to treebank construction rely heavily on manual annotation by linguistic experts, making the process expensive and time-consuming \autocite[233]{xuePennChineseTreeBank2005}.

  Recent work has explored various strategies for reducing annotation costs while maintaining quality. These include projection methods that transfer annotations across languages \autocite{hwaEvaluatingTranslationalCorrespondence2002}; bootstrapping approaches that iteratively improve annotations \autocite{mccloskyEffectiveSelftrainingParsing2006}; and cross-lingual transfer techniques that leverage existing resources \autocite{mcdonaldMultisourceTransferDelexicalized2011}. The Universal Dependencies project has provided a standardized framework for cross-lingual treebank development, facilitating comparison and transfer across languages \autocite{nivreUniversalDependenciesV12016,nivreUniversalDependenciesV22020}.

  For historical languages, treebank adaptation faces additional challenges due to the linguistic distance between historical and modern varieties. \textcite{eckhoffPROIELTreebankFamily2018} document a family of treebanks spanning several early Indo-European languages under a common annotation standard, while \textcite{zeldesGUMCorpusCreating2017} examined techniques for handling the specific annotation challenges posed by multilayer corpora.

  The work presented in this paper builds on these foundations by proposing an iterative human-in-the-loop procedure tailored to corpora situated at an extreme distance from available resources. Unlike approaches that emphasize transfer from related languages or domains, the method presented here systematically integrates curated human corrections into a bootstrapping loop that yields rapid gains on both morphological and syntactic tasks for a specialized documentary corpus.

  \section{The DALME-Marseille Corpus}

  The methodology presented in the paper is illustrated by a case study of a specialized collection of late-medieval inventories whose format and language make automatic processing highly challenging. This section characterizes the documentary context of these sources and the social and juridical milieux that produced them, as well as the linguistic properties that render them both analytically valuable and computationally difficult.

  The corpus is drawn from the \emph{Documentary Archaeology of Late Medieval Europe} (DALME) project, which investigates the material culture of later medieval Europe through written sources \autocite{smailDALMEDocumentaryArchaeology2020}. DALME makes inventories and related lists of objects accessible as open, well-structured, and machine-actionable datasets for computational analysis to enable a range of historical enquiries, from detailed case studies to large-scale quantitative research \autocite{pizzornoArchaeologyTextualThings2018}.

  A central challenge for this work is that comparable objects are recorded across Europe in many languages and dialects, each with their own vocabularies for documenting material culture \autocite{smailOverview2020}. The project's methodological premise is that such documentary descriptions encode classificatory regimes. A description is produced when someone classifies an object according to a representational system (for example, a contemporary curator's formal ontology, or a medieval notary's folk taxonomy). Because these systems are consistent within a bounded context, their encoding grammars may be extrapolated from data and used to map divergent descriptions onto a common schema. This mapping is what renders otherwise incomparable datasets interoperable \autocite{pizzornoArchaeologyTextualThings2018}.

  Accordingly, DALME treats museum catalogues, archaeological databases, and notarial registers alike, as linguistic encodings of material things. By analysing them at scale, it seeks to recover the taxonomic conventions that guided their construction and to map those conventions onto shared ontological frameworks \autocite{pizzornoMethodologyDocumentaryArchaeology2020}. Morphological and syntactic annotation of the type addressed in this paper is therefore integral to the project rather than ancillary.

  \subsection{Coverage and Provenance}

  The dataset analyzed here, referred to as \emph{DALME-Marseille}, comprises 160 inventories produced in the city of Marseille between 1258 and 1446. The collection constitutes an early longitudinal sample of domestic and institutional material culture that is unusually dense and socially inclusive for Western Europe. It spans the period during which the diffusion of paper broadened notarial record-keeping and encompasses the demographic and economic upheavals associated with the Black Death \autocite{smailPersonsThingsMarseille2020}.

  The documents derive from a variety of archival series and legal processes. Inventories occur in court registers, the records of hospitals and religious foundations, and notarial volumes concerned with estate administration and household auctions. They arise in a broad range of juridical contexts, including post-mortem estate surveys, guardianship proceedings, insolvency and confiscation cases, dowry and donation acts, and the acceptance of inheritances \autocite{smailGreaterProvence2024}. Consequently, the corpus captures routine aspects of late-medieval property law as it intersected with household goods, debts, and marital property.

  Socially, the corpus spans the urban hierarchy, from noble and large mercantile households, to artisans and labourers, to marginalized or impoverished actors, as well as ecclesiastical institutions and clerical households \autocite{smailFacePoverty2021,mcdonoughBrothel2021,smailPriestsInventory2022,smailCosmopolitanFisherman2023}. Women appear both as widows and guardians and as independent property holders and commercial agents. A subset of inventories documents Jewish households and others explicitly mention enslaved individuals \autocite{lowJewishHouseholds2023,smailWomanMerchantResens2021}. The result is a socially heterogeneous but structurally comparable corpus that reflects the material environments of diverse social, economic, and religious groups across nearly two centuries.

  \subsection{Linguistic Features}

  From a linguistic point of view, the Marseille inventories differ markedly from the theological, legal-theoretical, and literary Latin that dominate most historical corpora. They are working documents whose language was shaped by the pragmatic necessity of describing objects unambiguously for administrative and legal purposes. This produces a hybrid lexicon in which ordinary Medieval Latin coexists with specialized technical terms, locally used Occitan words, and formulaic legal expressions tied to particular procedures.

  Within the spatial and temporal bounds of the corpus, vocabulary for material culture is both stable and tightly circumscribed \autocite{pizzornoArchaeologyTextualThings2018}. Over nearly two centuries, notaries reuse similar phraseology and syntactic patterns to describe objects and their relations, indicating reliance on shared professional conventions and a common classificatory scheme. Scribes frequently juxtapose Latin descriptions with vernacular glosses, providing bilingual cues within individual phrases.

  The inventories vary considerably in length, with some listing as few as six items, while others enumerate three hundred or more. The complete dataset comprises 47,391 words organized into 6,903 sentences.

  The documents exhibit a strongly patterned macrostructure. Most acts open with conventional formulae setting out date, place, participants, and the legal basis for the inventory. They proceed through spatially ordered (often room-by-room) listings of goods and close with formulaic attestations, valuations, and signatures. The listings follow consistent syntactic templates for enumerating objects, quantities, and attributes, frequently employing markers that denote transitions between sections or categories. These characteristics—regularity and a constrained domain of discourse—distinguish the corpus as a good target for computational methods that exploit structural cues, yet they are also central to what makes automated processing difficult.

  \subsection{Challenges for Automated Processing}
  \label{sec:challenges}

  While the challenges that any automated processing of the DALME-Marseille corpus would present are multifaceted, they can be condensed into four concrete obstacles, each of which is shown to be consequential by the baseline evaluation in Section~\ref{sec:baselines}.

  \subsubsection{Multilingual Code-Switching}

  In the DALME-Marseille corpus, Latin and Occitan elements intermix within documents and within single phrases. In the 1,241 non-punctuation words of the gold standard (see Section~\ref{sec:gold}), 8.7\% are identifiable as Occitan and 88.6\% as Latin, with a further 2.6\% consisting of proper names that carry no independent language marking (only one token resists classification entirely). More strikingly, 36.5\% of sentences contain material from both languages. This degree of mixing is structural rather than incidental and contravenes the monolingual assumption underlying standard processing pipelines \autocite{cetinogluChallengesComputationalProcessing2016}.

  This lingual mixing is not evenly distributed across the vocabulary. Vernacular items cluster in the names of objects, materials, and locally specific artefacts—the domains where Latin lacked suitable equivalents—while the surrounding grammatical apparatus (prepositions, quantifiers, enumerative frames) remains Latin. The result is a hybrid configuration in which vernacular lexis is embedded in Latin morphosyntax, a pattern well attested in the Latin sociolinguistic literature \autocites[399-403]{adamsBilingualismLatinLanguage2003}[more generally, see:][]{wrightSociophilologicalStudyLate2002}{papaconstantinouMultilingualismGraecoRomanWorlds2012} and known to challenge parsers trained on monolingual data \autocite{bhatUniversalDependencyParsing2018}.

  \subsubsection{Ellipsis and the Enumerative Frame}

  Inventory syntax in the corpus departs markedly from the literary and religious Latin present in most treebanks \autocite{bammanAncientGreekLatin2011,mcgillivrayToolsHistoricalCorpus2013}. The most salient difference is the prevalence of ellipsis. Verbs are scarce, with nearly nine sentences in ten (88.8\%) beginning with the enumerative marker \emph{item}, and 84.3\% of those sentences containing no verb. Descriptive phrases often omit the copula (\emph{e.g.} \emph{tunica rubea} instead of \emph{tunica est rubea}), and determiners are frequently elided within formulaic frames. Dependency parsers trained on prose typically expect a predicate around which a clause is organized but such a predicate is often absent in these sentences.

  \subsubsection{Formulaic Repetition}

  Formulaic repetition with lexical substitution compounds the effect of ellipsis. The corpus contains 2,287 distinct part-of-speech (PoS) sequences amongst 6,903 sentences, and the ten most common patterns account for 44.1\% of all sentences. The single most common pattern—an enumerative marker followed by a numeral and a noun—occurs in 1,439 sentences. Such regularity offers an opportunity as well as a challenge. While a small annotated sample can be highly informative, statistical parsers are forced to generalize across structurally identical frames while accommodating the variable domain vocabulary that appears within them \autocite{echelmeyerPoStaggerFurMittelhochdeutsche2017}.

  \subsubsection{Orthographic Instability and the Lexicon}

  The corpus is rich in specialized terminology for material culture, legal procedure, and administrative practice that is poorly represented in standard Latin corpora. Some terms occur in medieval glossaries but not in standard dictionaries, while others reflect local usage absent from any lexicographic source \autocite{ducangeGlossariumMediaeInfimae1840,stotzHandbuchZurLateinischen1996}.

  The practical consequence is quantifiable. Relative to the vocabulary of the largest available Latin treebank, 68\% of the distinct lemmata in the DALME-Marseille gold standard are out-of-vocabulary. The affected terms include object names and attributes, regionally varying units of measurement and valuation, and folk-taxonomic descriptors whose distinctions were meaningful to medieval users but are opaque to systems trained on literary Latin \autocite{piotrowskiNaturalLanguageProcessing2012,bollmannLargescaleComparisonHistorical2019}.

  \begin{table*}[t]
 \centering
 \sffamily\footnotesize
 \begin{tabularx}{\textwidth}{Xllrrrrr}
  \toprule
  \scriptsize\textbf{Treebank} & \scriptsize\textbf{Genre} & \scriptsize\textbf{Temp. Coverage} & \scriptsize\textbf{Sentences} & \scriptsize\textbf{Tokens} & \scriptsize\textbf{Words} & \scriptsize\textbf{Unique Lem.} & \scriptsize\textbf{Words/sent.} \\
  \midrule
  Perseus & literary & 63 BC--AD 382 & 2,273 & 29,574 & 25,067 & 4,611 & 11.03 \\
  PROIEL & literary & 58 BC--AD 450 & 18,689 & 205,566 & 205,566 & 8,583 & 11.0 \\
  LLCT & documentary & AD 774--897 & 9,023 & 242,431 & 207,166 & 3,489 & 22.96 \\
  ITTB & literary & AD 1256--1274 & 26,977 & 450,554 & 388,145 & 5,209 & 14.39 \\
  UDante & literary & AD 1283--1320 & 1,723 & 55,818 & 47,206 & 5,619 & 27.4 \\
  DALME-Marseille & documentary & AD 1258--1446 & 6,903 & 54,294 & 47,391 & 3,499 & 6.87 \\
  \bottomrule
  \end{tabularx}
  \caption{Corpus statistics for existing treebanks. Unique lemmata are counted after case-folding and stripping digits and punctuation, so that variants of one lemma are not counted separately. For DALME-Marseille this gives 3,499 types against 3,585 distinct lemma strings.}
  \label{tab:treebank-stats}  
\end{table*}

  \paragraph{Orthographic instability defeats lexicon lookup.} In the absence of a standardized orthography, scribal spelling varies in two principal ways \autocites[48]{hermanVulgarLatin2000}[see also:][]{papaconstantinouMultilingualismGraecoRomanWorlds2012, hemelrijkRecentStudiesWord2006, adamsEarlyLateLatin2016, bollmannRulebasedNormalizationHistorical2011}. First, some words alternate within the corpus, for example simplified doubled consonants (\emph{quatuor} for \emph{quattuor}), or unsystematic uses of the latter (\emph{barrile} besides \emph{barrille}), together with other idiosyncratic forms (\emph{condam} occurs alongside \emph{quondam} at a ratio of 65 to 27). Second, other items exhibit stable non-classical forms so that no alternation is visible. Classical diphthongs reduce (\emph{oe}, \emph{ae} to \emph{e}), consonant developments reflect post-classical phonology (\emph{ti}/\emph{ci}), and the letter \emph{y}, originally associated with Greek loans, appears as a hypercorrection in tokens such as \emph{ymago} (ten instances) with no occurrences of classical \emph{imago}. Morphological regularization compounds both, favouring transparent formations over irregular classical alternatives (\emph{auricula} for \emph{auris}, \emph{agnellus} for \emph{agnus}).

  The computational consequence is direct and severe, since each orthographic variant constitutes a distinct surface form, lemmatizers that rely on exact-form matching fail on all but the single spelling they cover. Furthermore, the variants cannot be naively normalized away since some encode register or semantic distinctions that are analytically relevant, and any normalization would require an explicit target standard that does not exist for this material.

  \paragraph{The lexicon is conservative at its core and innovative at its edges.} As far as its core vocabulary is concerned, administrative Latin resists the changes associated with the Romance transition. For example, \emph{h}-deletion is rare (\emph{hospicium} is generally preferred to \emph{ospicium}), and terms such as \emph{domus} and \emph{vetus} persist, where vernacular alternatives might be expected. Innovation is concentrated in lexical domains central to the corpus's function—object names, materials, and measures—where Classical Latin lacked suitable terms and scribes adopted vernacular alternatives. This division explains an ostensibly paradoxical feature of the baseline results. The grammatical skeleton of the texts is conservative enough that models trained on literary Latin can often tag function words and recognize case morphology satisfactorily, but fail to properly process the lexical content borne by those structures.

  \section{Baseline Performance Analysis}
  \label{sec:baselines}

  In order to establish the need for a specialized adaptation procedure, off-the-shelf Latin models were evaluated on the DALME-Marseille gold standard. This baseline quantified the performance gap between widely used resources and the requirements of the specialized target corpus, and provided a reference against which to measure the gains achieved by the human-in-the-loop methodology.

  The evaluation targeted the core annotation tasks that underpin higher-level analyzes, namely universal and language-specific PoS tagging, morphological feature prediction, lemmatization, and dependency parsing, since deficiencies on these tasks would directly limit downstream capabilities such as information extraction, syntactic querying, and semantic analysis.

  \begin{table*}[t]
 \centering
 \sffamily\footnotesize
 \begin{tabularx}{\textwidth}{X|cccccc|ccccc}
  \toprule
   & \multicolumn{6}{c|}{\scriptsize\textbf{Morphological}} & \multicolumn{5}{c}{\scriptsize\textbf{Syntactic}} \\
  \cmidrule{2-12}
  \scriptsize\textbf{Model} & \scriptsize\textbf{~UPoS~} & \scriptsize\textbf{~XPoS~} & \scriptsize\textbf{Features} & \scriptsize\textbf{Lemmata} & \scriptsize\textbf{Ensemble} & \scriptsize\textbf{All Tags} & \scriptsize\textbf{~UAS~} & \scriptsize\textbf{~LAS~} & \scriptsize\textbf{~CLAS~} & \scriptsize\textbf{~MLAS~} & \scriptsize\textbf{~BLEX~} \\
  \midrule
  ITTB & .80 & .66 & .64 & .59 & .78 & .57 & .65 & .48 & .43 & .17 & .26 \\
  LLCT & .85 & .55 & .53 & .55 & .79 & .49 & .73 & .62 & .59 & .22 & .26 \\
  Perseus & .76 & .51 & .57 & .76 & .93 & .47 & .58 & .41 & .39 & .15 & .29 \\
  PROIEL & .73 & .53 & .66 & .66 & .79 & .49 & .43 & .28 & .27 & .15 & .22 \\
  UDante & .85 & .64 & .55 & .71 & .91 & .52 & .70 & .57 & .52 & .24 & .37 \\
  \bottomrule
  \end{tabularx}
  \caption{Performance of models trained on existing Latin treebanks, applied to the 200-sentence DALME-Marseille gold standard. Metrics are discussed in Section~\ref{sec:metrics}.}
  \label{tab:pre-trained-performance}  
\end{table*}

  \subsection{Models and Treebanks}

  The models selected for the baseline were trained on five Latin treebanks available in the Universal Dependencies (UD) format: the Index Thomisticus Treebank (ITTB)\footnote{The Index comprises the complete works of Thomas Aquinas and related scholastic authors \autocite{cecchiniChallengesConvertingIndex2018,passarottiImprovementsParsingIndex2010}.}, the Late Latin Charter Treebank (LLCT)\footnote{LLCT is drawn from early medieval charters \autocite{cecchiniNewLatinTreebank2020,korkiakangasLateLatinCharter2021}.}, the Perseus UD Latin treebank\footnote{The Perseus treebank is derived from a selection of passages from the Ancient Greek and Latin Dependency Treebank \autocite{bammanAncientGreekLatin2011}.}, the PROIEL treebank\footnote{PROIEL includes most of the Vulgate New Testament translations plus selections from a number of classical texts \autocite{haugCreatingParallelTreebank2008}.}, and UDante\footnote{UDante is based on Latin texts attributed to Dante Alighieri \autocite{cecchiniUDanteFirstSteps2020}.} (Table~\ref{tab:treebank-stats}).

  These treebanks differ in period, genre, and register. ITTB represents medieval scholastic Latin, LLCT includes documentary and early medieval Latin, Perseus is classical literary Latin, UDante a single fourteenth-century author, and PROIEL combines biblical and classical material. Evaluating models trained on this set allowed us to assess whether any existing variety of Latin provided adequate coverage for the DALME inventory language.\footnote{CIRCSE (\url{https://universaldependencies.org/treebanks/la_circse/index.html}) is a further UD Latin treebank, but Stanza distributes no pre-trained model for it, so it was not included in the model-level comparisons.}

  Each model was evaluated using the Stanza NLP framework \autocite{qiStanzaPythonNatural2020}, on the same set of 200 manually annotated sentences from DALME-Marseille  (see Section~\ref{sec:gold}).\footnote{The Stanza version used was 1.2.1, the latest release when the experiment began. Section~\ref{sec:modern-baseline} discusses the current release's performance on the same data.}

  \subsection{Evaluation Metrics}
  \label{sec:metrics}

  Evaluation throughout this paper uses the CoNLL 2018 shared-task metric set \autocite[4-6]{zemanCoNLL2018Shared2018}.

  Morphological metrics measure performance when tagging universal part-of-speech (UPoS), language-specific part-of-speech (XPoS), morphological features (Features), and lemma (Lemmata). \emph{All Tags} requires those four to be simultaneously correct on a token, while an additional morphological metric, the \emph{Ensemble} lemmatization score, is described below (Section~\ref{sec:lexicon-rebuild}).

  Syntactic metrics measure performance when mapping unlabelled (UAS) and labelled attachment (LAS), content-word labelled attachment (CLAS, which scores only relations between content-words), morphology-aware labelled attachment (MLAS, which requires attachment, label, and morphological analysis to be correct together), and bi-lexical dependency (BLEX, which requires attachment, label, and lemma to be correct) \autocite[6-7]{nivre-fang-2017-universal}. MLAS is the most stringent of the syntactic metrics and the one on which models diverge most. All reported values are F1.

  \begin{figure*}[t]
    \centering
    \includegraphics[alt={Scatter plot placing the Latin models in context. Grey points are the 100 Stanza models released for Universal Dependencies 2.5, scored on their own test sets; arrows show how far each Latin model falls when evaluated instead on the DALME-Marseille gold standard.}, width=\textwidth]{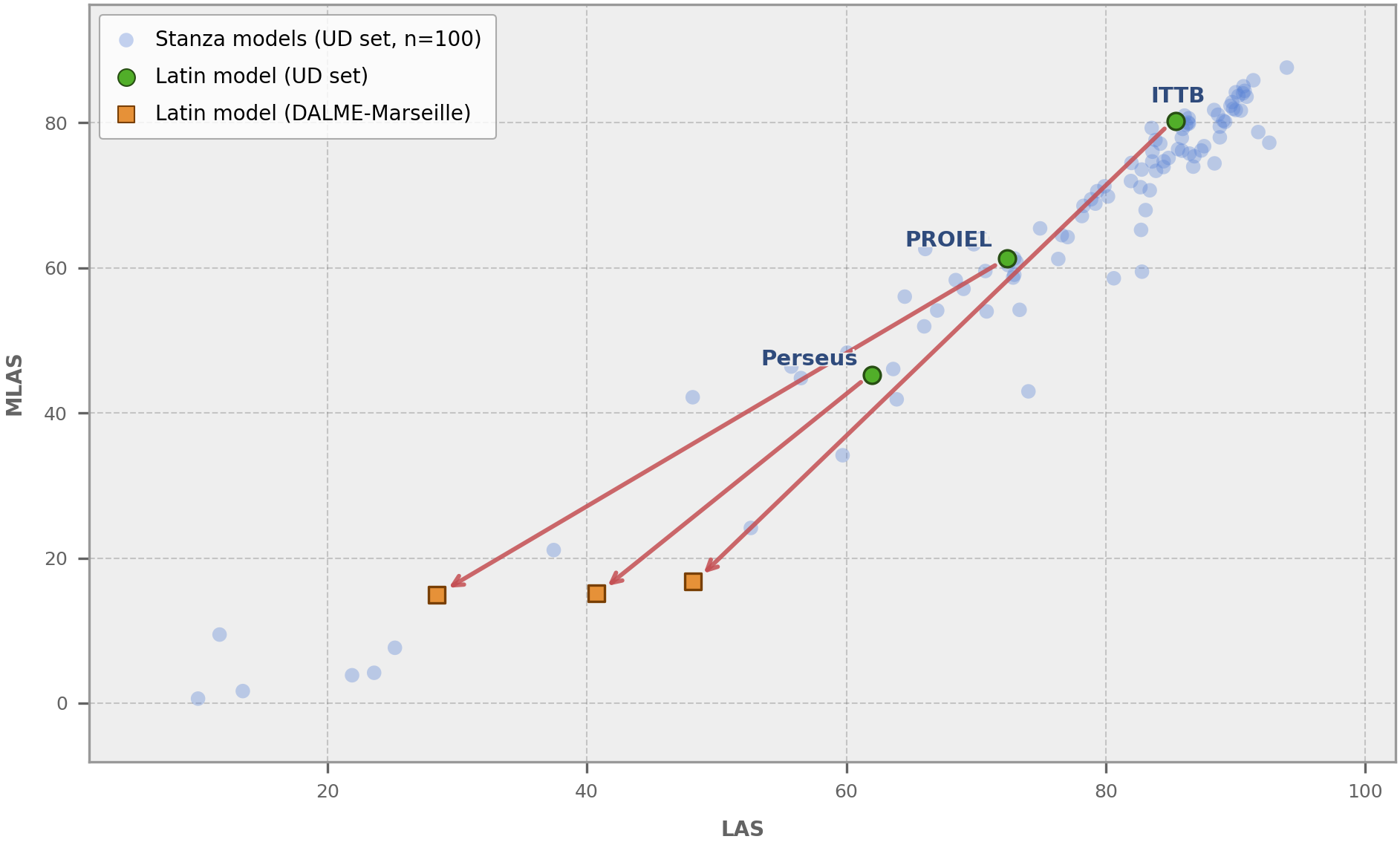}
    \caption{Relative performance of Latin models across domains. Grey points show the 100 Stanza models released for UD 2.5 on their native test sets. Arrows indicate the score change when the same model is evaluated on the DALME-Marseille gold standard. ITTB occupies the 84th percentile on UPoS and the 81st on MLAS amongst the released models, yet it loses 63.5 MLAS points when applied to the target corpus.}
    \label{fig:latin-in-context}  
  \end{figure*}

  \subsection{Baseline Results}

  The results show that while some models attain acceptable performance on individual tasks, none performs adequately across the full annotation pipeline (Table~\ref{tab:pre-trained-performance}). LLCT and UDante each achieved 85\% on UPoS tagging, and Perseus reached 76\% on lemmatization, but the best All Tags score was only 57\% (ITTB), indicating that fewer than three tokens in five received a fully correct morphological analysis. These figures lie well below conventional thresholds for reliable downstream use (≈90\%) \autocite[171]{manningPartofSpeechTagging972011}.

  Lemmatization was particularly poor (55–76\% for sequence-to-sequence lemmatizers), reflecting the lexical challenges discussed in Section~\ref{sec:challenges}. The prevalence of vernacular vocabulary, technical terms, and orthographic variation produces systematic out-of-vocabulary failure modes for models trained on standard Latin treebanks \autocite{bollmannLargescaleComparisonHistorical2019,piotrowskiNaturalLanguageProcessing2012}.

  To mitigate lexical gaps, an ensemble lemmatization approach was implemented that prioritizes dictionary lookup from a lexicon. When a surface form appears in the lexicon it is resolved by lookup and only unknown forms are passed to the sequence-to-sequence lemmatizer. This intervention raised lemmatization by 13–24 percentage points depending on the base model (Table~\ref{tab:pre-trained-performance}, \emph{Ensemble}), with the largest absolute gains going to the models having the weakest native lexical coverage. The improvement is therefore a property of lexicon coverage rather than of the underlying models, and even with the ensemble every configuration remained far from usable syntactic accuracy.

  No single pre-trained package dominated across tasks. ITTB led on XPoS, Features, and All Tags. LLCT and UDante shared the best UPoS score, and Perseus attained the strongest pure lemmatizer score, both with and without the ensemble. LLCT generally led on syntactic metrics, except BLEX and MLAS, where UDante was ahead. Critically, the best LAS was 0.62 and the best MLAS was only 0.24, levels that are insufficient for the syntactic querying for which the corpus is intended.

  Figure~\ref{fig:latin-in-context} places these results in the broader context of all Stanza-released models. The Latin baselines perform well on their native test sets—ITTB sits in the top fifth on the morphology-aware metric—but degrade dramatically on the DALME inventory language, quantifying the domain effect targeted by this study.

  \begin{table*}[t]
 \centering
 \sffamily\footnotesize
 \begin{tabularx}{\textwidth}{XXccccccccc}
  \toprule
  \scriptsize\textbf{Treebank} & \scriptsize\textbf{Stanza} & \scriptsize\textbf{UPoS} & \scriptsize\textbf{XPoS} & \scriptsize\textbf{Features} & \scriptsize\textbf{Lemmata} & \scriptsize\textbf{All Tags} & \scriptsize\textbf{UAS} & \scriptsize\textbf{LAS} & \scriptsize\textbf{MLAS} & \scriptsize\textbf{BLEX} \\
  \midrule
  ITTB & 1.2.1 & .80 & .66 & .64 & .59 & .57 & .65 & .48 & .17 & .26 \\
   & 1.14 & .72 & .58 & .57 & .74 & .53 & .62 & .44 & .18 & .28 \\
  \midrule
  LLCT & 1.2.1 & .85 & .55 & .53 & .55 & .49 & .73 & .62 & .22 & .26 \\
   & 1.14 & .77 & .51 & .52 & .57 & .48 & .72 & .54 & .21 & .18 \\
  \midrule
  Perseus & 1.2.1 & .76 & .51 & .57 & .76 & .47 & .58 & .41 & .15 & .29 \\
   & 1.14 & .76 & .53 & .54 & .76 & .48 & .52 & .37 & .12 & .24 \\
  \midrule
  PROIEL & 1.2.1 & .73 & .53 & .66 & .66 & .49 & .43 & .28 & .15 & .22 \\
   & 1.14 & .72 & .53 & .54 & .65 & .49 & .43 & .33 & .18 & .26 \\
  \midrule
  UDante & 1.2.1 & .85 & .64 & .55 & .71 & .52 & .70 & .57 & .24 & .37 \\
   & 1.14 & .74 & .54 & .55 & .71 & .50 & .59 & .40 & .15 & .23 \\
  \bottomrule
  \end{tabularx}
  \caption{Current off-the-shelf Stanza Latin models on the same gold standard versus version 1.2.1. Neither row uses the domain lexicon, so both are directly comparable.}
  \label{tab:modern-stanza}  
\end{table*}

  \subsection{Lexicon and Evaluation Independence}
  \label{sec:lexicon-rebuild}

  Because the lexicon was derived from the corpus, and the gold standard is a subset of that corpus, naive inclusion of gold-only lemmas would inflate ensemble scores through contamination. To prevent this, every entry attested \emph{only} in gold-standard sentences (61 of 5,315 entries) was removed from the lexicon.

  The correction has a measurable effect, with lexicon coverage of gold-standard surface forms falling from 96.9\% to 91.5\%, and the proportion of gold-standard words whose exact lemma the lexicon supplies by lookup falling from 95.5\% to 90.2\%. All ensemble figures reported in this paper use the reduced lexicon. For comparison, the uncorrected lexicon would raise the \emph{Ensemble} column by 2.4 to 3.8 points and BLEX by 0.8 to 3.7 points; no other metric would move by more than 0.1.

  \subsection{Does a Newer Release Close the Gap?}
  \label{sec:modern-baseline}

  The current Stanza release (1.14) was also evaluated on the same gold standard, using identical pre-tokenized input and no domain lexicon, to test whether improvements in the distributed packages had reduced the domain gap (Table~\ref{tab:modern-stanza}).

  The gap remains. Four of five packages lose UPoS accuracy (UDante by 11 points), and four of five score lower on LAS (UDante by 17 points). The exceptions are limited, with PROIEL gaining ≈5 points on LAS and ITTB increasing lemmatization scores by 15 points. The best out-of-the-box result from the current release is LLCT at 0.77 UPoS and 0.54 LAS, compared with 0.85 and 0.62 for the earlier package.

  These results should not be interpreted as framework regression, since model packages differ between releases and are trained on different treebank snapshots. The comparison does indicate, however, that general-purpose Latin packages as distributed remain substantially inadequate for this documentary inventory language. The empirical case for domain-specific adaptation is therefore at least as strong today as when the experiments began.

  It is important to note that this comparison concerns packages \emph{as shipped}, and current Stanza ships no transformer-backed Latin package—only the variants tested above. Latin-specific transformer models now exist that can be supplied to the same training pipeline as a backbone, and a system so built could potentially attain better results. Such an experiment is left for future work (see Section~\ref{sec:future}).

  \subsection{Genre and Period Do Not Predict Transfer}
  \label{sec:genre-period}

  The results also indicate that proximity in genre or period is a poor predictor of transfer. Although LLCT is the closest corpus by genre (documentary Latin), and is the best baseline on LAS, its morphological performance was uneven, leading on UPoS but ranking last on features (0.528) and on lemmatization without the ensemble (0.554). UDante, comprising a single fourteenth-century literary author, outperformed LLCT on seven of eleven metrics.

  The practical implication is that lexicon overlap and structural profile—both computable from an unannotated sample—are more informative selection criteria for a starting model than simple period or genre proximity.

  \subsection{Corpus Comparison and Analysis}
  \label{sec:corpus-comparison}

  As a means to clarify the baseline failures, the DALME-Marseille corpus was systematically compared to the five treebanks whose models were included in the evaluation.

  As expected, the corpus exhibits features characteristic of late Medieval Latin, with word order settling toward vernacular patterns, prepositions used to reinforce case relations, \emph{quod}/\emph{quia}-clauses replacing accusative/infinitive, \emph{habeo}/\emph{debeo} used for the gerundive of obligation, and demonstratives drifting towards article-like function \autocite{hemelrijkRecentStudiesWord2006,adamsSocialVariationLatin2013,clacksonBlackwellHistoryLatin2007,hermanVulgarLatin2000}. In terms of characteristics most relevant to model transfer, they can be reduced to three measurable dimensions: sentence length, dependency structure, and morphological composition.

  \subsubsection{Structural and Syntactic Complexity}

  The most salient difference between DALME-Marseille and the comparison Latin treebanks concerns their fundamental structural characteristics. Sentence length, defined as the number of words per sentence, serves as a proxy for syntactic complexity and exposes marked contrasts between documentary inventories and literary or theological texts.

  DALME-Marseille has a mean sentence length of 6.87 words, substantially shorter than those of the comparison treebanks. Literary treebanks exhibit approximately double or greater mean lengths, with Perseus averaging 11.03 words per sentence, PROIEL 11.00, ITTB 14.39, and UDante 27.40.

  The observed disparity in sentence length reflects substantive differences in genre and purpose. Documents in the DALME-Marseille corpus adopt highly formulaic enumerative frames designed for administrative clarity and legal precision, rather than rhetorical elaboration. By contrast, the literary and theological texts represented in the comparison treebanks make extensive use of subordinate constructions, rhetorical devices, and extended argumentation, producing substantially greater syntactic complexity \autocite{biberRegisterGenreStyle2009,leeGenresRegistersText2001}.

  \begin{table}[H]
 \centering
 \sffamily\footnotesize
 \begin{tabulary}{\textwidth}{LRRRR}
  \toprule
  \scriptsize\textbf{Corpus} & \scriptsize\textbf{Sents.} & \scriptsize\textbf{r} & \scriptsize\textbf{\small{$R^2$}} & \scriptsize\textbf{Mean len.} \\
  \midrule
  DALME-Marseille & 6,894 & .848 & .720 & 6.87 \\
  PROIEL & 18,412 & .821 & .674 & 11.15 \\
  Perseus & 2,262 & .784 & .615 & 11.08 \\
  ITTB & 25,302 & .761 & .579 & 15.27 \\
  LLCT & 9,016 & .718 & .516 & 22.98 \\
  UDante & 1,723 & .677 & .459 & 27.40 \\
  \bottomrule
  \end{tabulary}
  \caption{Pearson correlation between sentence length in words and dependency tree depth. All correlations are significant at $p < 10^{-200}$. Sentences of fewer than two non-punctuation words are excluded, which is why the mean lengths differ slightly from those in Table~\ref{tab:treebank-stats}.}
  \label{tab:length-correlations}  
\end{table}

  \subsubsection{Length and Dependency Structure}

  Since sentence length may appear as a coarse proxy for syntactic complexity, its relationship to dependency-tree depth was also measured. A tight relationship would imply that sentences of a given length have a predictable shape, while a loose relationship would indicate that identical lengths can correspond to structurally diverse sentences (Table~\ref{tab:length-correlations}).

  DALME-Marseille exhibits the strongest coupling amongst the six corpora ($r=0.848$, $R^2=0.720$). The corpora order nearly monotonically by mean sentence length, so shorter sentences tend to have more predictable structure. This property is expected for formulaic enumeration, and is precisely what the bootstrapping method exploits, given that a small annotated sample can cover a large proportion of the structural variety actually present.

  \begin{figure*}[t]
    \centering
    \includegraphics[alt={Stacked horizontal bars comparing part-of-speech group shares across the six corpora as a percentage of words, showing DALME-Marseille as markedly more nominal than the literary treebanks.}, width=\textwidth]{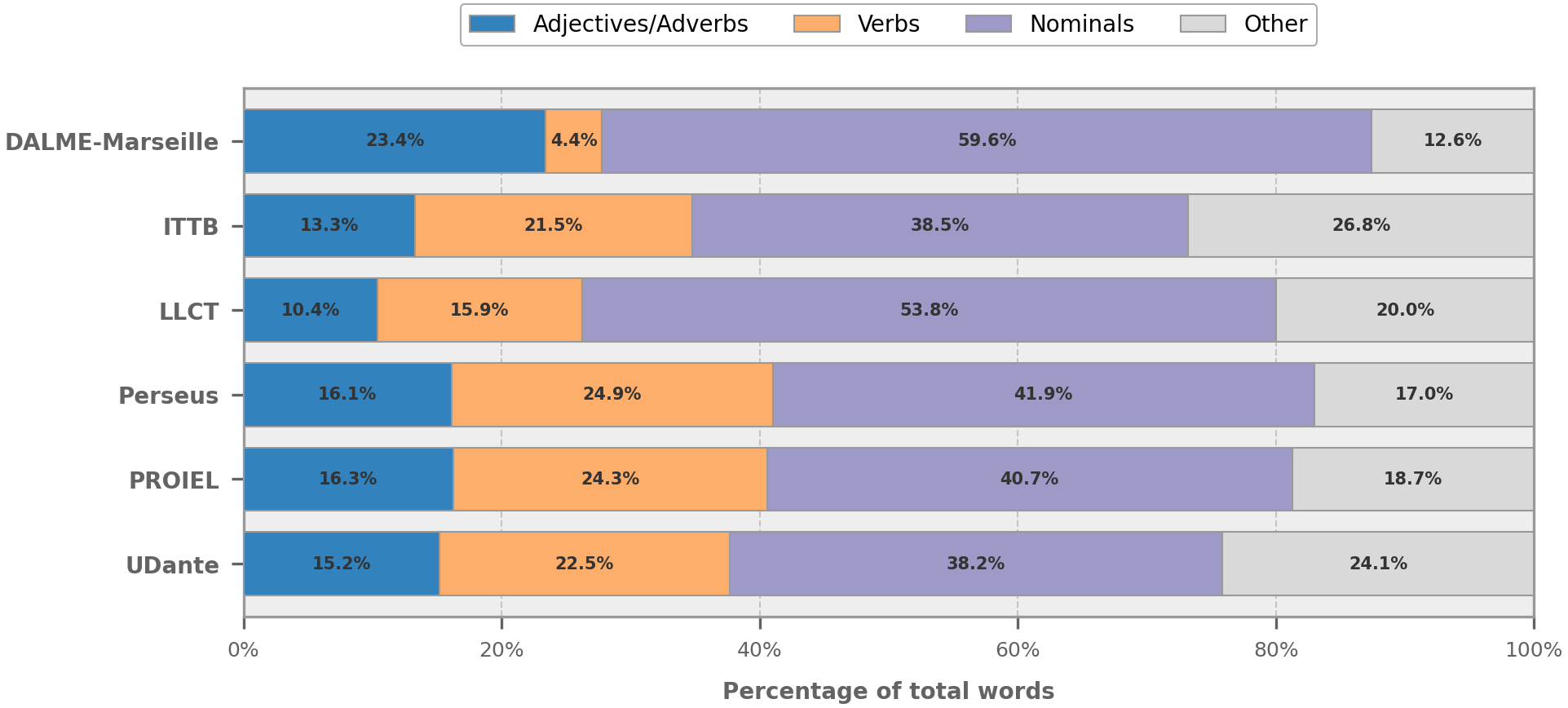}
    \caption{Part-of-speech group shares across the six corpora, as a percentage of words. Punctuation and the underspecified tags are excluded from the denominator as described in Section~\ref{sec:morpho-distribution}. DALME-Marseille has the largest nominal share and by far the smallest verbal one, 59.6\% and 4.4\% respectively, where no other corpus falls below 15.9\% verbs.}
    \label{fig:pos-distribution} 
  \end{figure*}

  The difference is one of degree rather than of kind. DALME-Marseille sits at one end of a continuum (0.848 versus PROIEL's 0.821), and the range between the most and least predictable corpora (0.848 versus 0.677) is modest. The corpus's distinguishing feature is therefore the extremity of its length distribution, rather than a qualitatively different length–depth relationship.

  A second measure supports the same conclusion and is more directly comparable with prior work \autocite[see, for example:][5]{oyaSyntacticDependencyDistance2011}.  Mean dependency distance in DALME-Marseille is 1.89 words, the smallest amongst the six corpora. The remaining means, ordered by sentence length, are PROIEL 2.20, Perseus 2.83, ITTB 3.00, LLCT 3.31, and UDante 3.93. Consequently, parsers applied to this material resolve attachments over shorter spans than on any of the comparison treebanks, which helps explain the rapid acquisition of the enumerative frame. Conversely, because short attachment spans amplify the impact of an incorrectly tagged function word, the early adposition failure discussed in Section~\ref{sec:adp-dip} is particularly damaging \autocites[28]{gibsonLinguisticComplexityLocality1998}[10339]{futrellLargescaleEvidenceDependency2015}.

  \subsubsection{Morphological Distribution Patterns}
  \label{sec:morpho-distribution}

  DALME-Marseille exhibits a substantially different morphological profile from existing Latin treebanks. Comparing part-of-speech distributions highlights grammatical differences that help explain the parsing difficulties reported in the baseline evaluations. The \emph{Nominals} category is defined to include nouns, proper nouns, pronouns, determiners, and numerals.\footnote{Determiners are grouped with pronouns because the boundary between the two is drawn differently in each treebank. PROIEL tags 3.7\% of words as determiners and 12.4\% as pronouns, LLCT 9.7\% and 8.8\%. Their sum is markedly more stable across the five (13.0–18.5\%) than either component alone, so grouping them removes a difference in annotation practice rather than one in the language. Numerals are included because, although they can function as determiners, adjectives, or pronouns, they are almost always determiners or pronouns in the DALME-Marseille corpus, where they account for 29.5\% of nominals against 0.8–2.0\% in the other five.} The \emph{Verbs} category aggregates main and auxiliary verbs, as the distinction is not directly relevant to the analysis and the latter category is used inconsistently across treebanks.\footnote{ITTB, Perseus, and UDante recognize only \emph{sum} as an auxiliary, LLCT adds \emph{habeo}, and PROIEL adds \emph{eo}. As a share of all verbal tokens, auxiliaries range from 6.0\% in Perseus to 28.3\% in ITTB, a spread that reflects annotation policy rather than the Latin of the texts.} Adjectives and adverbs are grouped in their own separate category, while interjections, particles, adpositions, and conjunctions are grouped under \emph{Other}.\footnote{Consistency is again the reason behind these categorizations: PROIEL tags the negators \emph{non} and \emph{ne} as adverbs, where ITTB, LLCT, Perseus, and UDante tag \emph{non} as a particle. The particle tag covers 5.7\% of ITTB but is entirely absent from PROIEL. The \emph{punctuation} and \emph{other} tags, PUNCT and X respectively, are excluded from the analysis, as they are not directly relevant and are not used consistently across treebanks. The symbol tag (SYM), which no Latin treebank uses, is likewise excluded.}

  As Figure~\ref{fig:pos-distribution} indicates, the part-of-speech profile of the DALME-Marseille corpus departs substantially from those of the comparison treebanks. The ratio of nominals to verbs is illustrative, ranging from 1.7:1 (Perseus) to 3.4:1 (LLCT) amongst the existing resources, but rising to 13.7:1 in DALME-Marseille.

  These proportions reflect the purpose of the documents. The function of the inventories is primarily to list and identify objects, producing a strong nominal bias. Predication is seldom required, so verbs account for only 3.75\% of tokens, with auxiliaries contributing an additional 0.60\%. Adjectives appear chiefly as descriptive modifiers, whereas the adverbial and conjunctive apparatus typical of literary prose is largely absent. The enumerative frame discussed in Section~\ref{sec:challenges} is the clearest example of this skew.

  This morphological bias presents concrete difficulties for NLP systems trained on more balanced corpora. Statistical models internalize expected distributional patterns, and the extreme nominal dominance observed here violates those expectations and contributes to the degraded parsing performance reported in the baseline evaluations. The sensitivity of parsers to genre-specific morphological distributions has been documented in prior work on domain adaptation \autocite{plankEffectiveMeasuresDomain2011,mccloskyAutomaticDomainAdaptation2010}.

  \subsection{Implications for Model Performance}

  The measurements above characterize the principal challenges confronting the baseline models. Two forces act in opposing directions. Structural regularity disadvantages a model that has not been exposed to this genre, but benefits one trained on in-domain data. The same formulaic frame that violates the distributional assumptions of a tagger trained on running prose is precisely what allows a few hundred annotated sentences to cover a large proportion of the structural variation in the DALME-Marseille corpus. Lexical innovation, by contrast, provides no analogous compensation. It is concentrated in the object vocabulary central to these documents, absent from every available treebank, and no quantity of additional out-of-domain Latin can substitute for it. In this configuration, conventional domain-adaptation techniques are weakest, because there is no appropriate in-domain corpus to adapt from, and the distance to available resources is too great for naive transfer to succeed \autocite[155]{ben-davidTheoryLearningDifferent2010}.

  The methodology presented below, therefore, exploits the tractable component (structural regularity) while acknowledging its limited reach on lexical gaps. This asymmetry explains which annotation categories improve through iteration, and which remain resistant (The results section below reports these outcomes separately).

  \section{Method}
  \label{sec:methodology}

  The previous section demonstrated that no existing Latin model performs adequately on this corpus. Furthermore, the deficit is primarily lexical and structural and cannot be addressed by the treebanks currently available. The procedure presented in this section responds directly to that diagnosis. In the absence of in-domain annotated data, such data must be produced, and the most cost-effective approach is to generate it as a by-product of editing model output rather than by annotating from scratch.

  The method follows a conventional allocation of labour, with the models providing speed and consistency on material they have encountered, and the human annotator supplying the judgements and domain vocabulary the models lack \autocite{settlesActiveLearning2012,amershiPowerPeopleRole2014}. Crucially, the annotator edits machine-produced analyzes rather than creating annotations \emph{de novo}. The efficiency gains derive from the diminishing size of these edits as models improve rather than from the selection of maximally informative examples.

  \subsection{Methodological Framework}

  The implementation consists of an iterative bootstrapping procedure where a model annotates a sampled batch, a human corrects the output, the corrections are added to the training pool, and a new model is trained on the accumulated data. The design is adjacent to self-training and incremental corpus construction in that each model's output contributes to subsequent training material, but differs in that every machine-produced annotation is inspected and corrected by an expert rather than filtered by automatic confidence measures \autocite{mccloskyEffectiveSelftrainingParsing2006}. The procedure is also motivated by the same cost-reduction objectives that underlie active learning \autocite{settlesActiveLearningLiterature2009}, although selection is held constant here (see Section~\ref{sec:sampling}).

  Four explicit design commitments govern the procedure: (1) \emph{Correction rather than annotation.} Instead of performing full manual annotation from first principles, the annotator edits machine-produced analyzes. This change of task reduces effort because the required edits decline as models improve (the reduction in token-level corrections over the chain is quantified in Section~\ref{sec:cost-benefit}). (2) \emph{Retraining rather than incremental fine-tuning.} Each iteration trains a fresh set of models on the entire accumulated pool instead of continuing previous model weights on the new batch. This approach increases computational cost and conflates the effects of additional data and of a fresh start, but it ensures that each model is a reproducible function of its training set alone, allowing direct comparisons across iterations (Section~\ref{sec:partitioning}) \autocite{tomanekSemisupervisedActiveLearning2009}. (3) \emph{Specialization over generality.} Models are specialized to this corpus. The training regime does not preserve out-of-domain performance, meaning that cross-corpus reusability is bound to be affected. For projects requiring multi-archive applicability, a different trade-off between specialization and generality would be necessary \autocite{panSurveyTransferLearning2010}. (4) \emph{Scheme stability.} A single annotator performed the corrections, so the risk to guard against is drift within one person, rather than disagreement between several. Section~\ref{sec:consistency} describes the measures used to monitor and mitigate drift.

  \begin{figure}[H]
    \centering
    \includegraphics[alt={Flow diagram of one iteration: sample a batch, pre-annotate it with the previous model, correct it by hand, pool it with earlier batches, and retrain. The loop closes from training back to sampling; evaluation records progress but does not feed back into selection.}, width=3in]{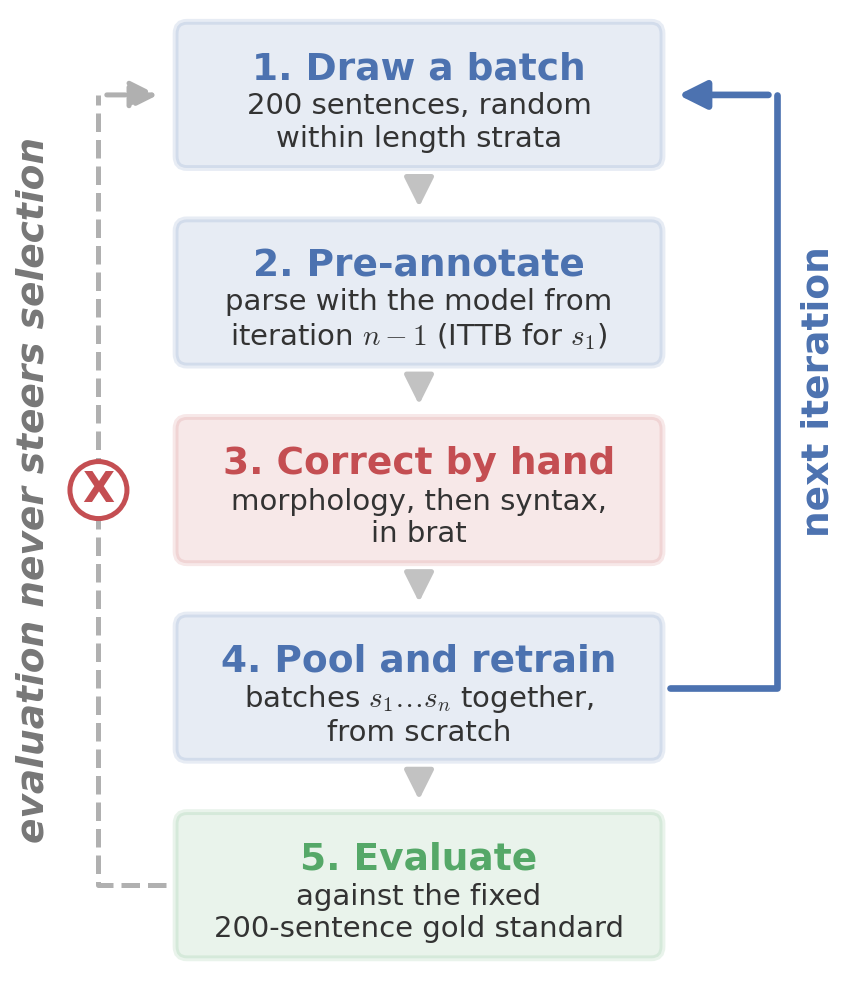}
    \caption{One iteration of the procedure. The loop closes from training back to sampling. Evaluation records progress but does not feed back into selection.}
    \label{fig:bootstrapping-loop}  
  \end{figure}

  One iteration comprises five fixed steps (Figure~\ref{fig:bootstrapping-loop}):

  \begin{enumerate}
    \item \textbf{Batch selection}: Draw 200 sentences at random within predefined length strata, excluding sentences selected in earlier rounds (Section~\ref{sec:sampling}).
    \item \textbf{Pre-annotation}: Annotate the batch automatically with the model from the previous iteration (the ITTB baseline provided the initial annotations for round one).
    \item \textbf{Manual correction}: Correct the morphological and syntactic analyzes by hand using \emph{brat}.
    \item \textbf{Training}: Pool the corrected batch with all prior corrected data, shuffle, partition into training/development/held-out portions (Section~\ref{sec:partitioning}), and retrain models from scratch.
    \item \textbf{Evaluation}: Evaluate the resulting models on the fixed gold standard (Section~\ref{sec:gold}).
  \end{enumerate}

  Crucially, evaluation does not influence selection. Step five records progress, but does not feed back into step one. The batch drawn for iteration $n+1$ is independent of the model produced at iteration $n$, which preserves the experimental contrast between adding in-domain data \emph{per se}, and any gains attributable to selective sampling.

  Maintaining identical procedures across iterations is both the experiment's principal methodological virtue and its greatest limitation. The uniformity yields an interpretable learning curve, attributable solely to the accumulation of in-domain data, but it precludes conclusions about whether alternative selection policies (for example, uncertainty—or diversity—based sampling) could have achieved equivalent accuracy with less annotation. That question is considered in Section~\ref{sec:future}.

  Section~\ref{sec:stopping} describes how the process was terminated in practice.

  \subsection{Corpus Preparation}

  Before sampling, the corpus exported from the project database underwent two deterministic preprocessing corrections. Roman numerals, which occur frequently in the material, were normalized in both surface form and lemma (2,910 forms, 1,504 distinct lemmata), and the \token{NumForm} feature was set consistently across 2,674 tokens.

  In addition, the lemma of the scribal variant \emph{condam} was regularized to \emph{quondam} (65 tokens). This regularization is distinct from orthographic normalization, which was deliberately \emph{not} performed. Only the lemma entries were harmonized, with all 65 surface occurrences of \emph{condam} remaining in the text alongside 27 occurrences of \emph{quondam}. This policy preserves surface spelling as evidence of scribal practice (Section~\ref{sec:challenges}) while unifying lemma forms so that variants can be counted together. The preprocessing steps are deterministic and released as reproducible code alongside the corpus.

  \subsection{Sampling and the Annotated Subset}
  \label{sec:sampling}

  Seed batches were drawn by random sampling stratified solely by sentence length. Sentences were assigned to three bins according to their count of non-punctuation words—short (fewer than 7), medium (7–12), and long (13 or more)—and each 200-sentence batch was composed to a fixed quota of 136 short, 46 medium, and 18 long sentences. Sampling was performed without replacement across rounds, so the nine annotated batches are disjoint. The quota approximates the corpus distribution (68.1\%, 22.9\%, 9.0\% respectively).

  Selection was not stratified by date, document type, social context, or lexical rarity, nor were any active-learning selection heuristics employed. Batches were drawn independently of model state, with sentences not chosen on the basis of model uncertainty, disagreement, or estimated information gain. This design is deliberately simple. Its advantage is methodological transparency—the learning curve in Section~\ref{sec:iteration-results} thus measures the effect of adding in-domain data, uncontaminated by a selection policy that might itself contribute to observed gains.

  The initial batch predates the regularized procedure and is a minor exception. It contains 204 sentences distributed 118/41/45 across the three bins. In aggregate, the annotated pool comprises 1,804 sentences and 14,516 tokens, representing 26\% of the corpus by sentence and 27\% by tokens.

  \subsection{Pre-annotation}

  Each sampled batch was pre-annotated automatically prior to manual correction. For rounds $s_2$ and later, the annotating model was the one produced by the previous iteration, while for the first round the ITTB baseline served as the initial pre-annotator because no domain-trained model yet existed. ITTB was selected because it was found to be the strongest baseline on the combined morphological measure (All Tags, 0.57), on XPoS, and on Features (Table~\ref{tab:pre-trained-performance}). Since the annotator's first pass corrects morphology before syntax, the model with the best morphological analysis is the one that leaves the least to correct, notwithstanding that LLCT parses better.

  Pre-annotation applied the full Stanza pipeline (tokenization, PoS tagging, morphological features, lemmatization, and dependency parsing). Lemmatization employed the ensemble configuration described in Section~\ref{sec:lexicon-rebuild}, whereby the domain lexicon is consulted prior to falling back to the sequence-to-sequence lemmatizer.

  A pragmatic preprocessing step materially affected lexical analysis. Entries in the domain lexicon were annotated with a language label—Latin, Occitan, or indeterminate—because a substantial vernacular component recurs in the corpus. This lexicon-level classification enables the code-switching analysis reported in Section~\ref{sec:oov}, and is applied at lookup time rather than inferred during parsing.

  The language classifier operates by dictionary lookup with a model fallback, so its coverage depends on the underlying lexical resources. For Latin those are the CLTK lemma lists \autocite{johnsonClassicalLanguageToolkit2021}, Lewis's \emph{Elementary Latin Dictionary} \autocite{lewisElementaryLatinDictionary1890}, and the lemma inventories of the five comparison treebanks. For Occitan, the UD-Occitan-TTB \autocite{miletic-etal-2020-four} and the \emph{Dictionnaire de l'Occitan Médiéval} \autocite{stempelDictionnaireLOccitanMedieval2012}. Lemmata absent from these sources are labelled as indeterminate, so the category reflects the coverage of the dictionaries rather than an absolute typological judgement. Medieval Occitan is substantially less well resourced than Latin and, accordingly, vernacular items are more likely to be unrecognized, contributing to the disproportionate vernacular component noted amongst residual out-of-vocabulary items.

  \subsection{Correction and Annotation Guidelines}

  Since correction constitutes the primary cost, and determines the dataset's quality, the annotation workflow and its procedural controls are described here in some detail.

  Manual correction was performed using the \emph{brat} annotation tool \autocite{stenetorpBratWebbasedTool2012}, a web-based interface that visualizes syntactic structure and permits direct editing of annotations\footnote{\url{https://brat.nlplab.org}}. For each batch, the pre-annotated output was converted to \emph{brat}'s standoff format at ten sentences per document, corrected in the web interface, and exported back to CoNLL-U.

  The correction workflow proceeded in two ordered passes. The first pass addressed morphology, with the annotator reviewing and correcting universal and language-specific PoS tags, morphological features, and lemma assignments generated by the pre-annotating model. These are local decisions, and resolving them first established the morphological constraints on subsequent syntactic analysis since the tag assigned to a token restricts the set of plausible dependency relations.

  The second pass addressed syntactic structure. Using \emph{brat}'s tree editor, the annotator corrected head assignments and dependency labels within the UD framework. This pass was structurally demanding, and required adjudication of non-local grammatical relationships.

  Throughout, the working principle was to prioritize consistency over theoretical maximalism. The formulaic and elliptical constructions characteristic of the corpus rarely admit a single uncontroversial analysis. The principal objective was therefore to ensure that identical structures received identical treatment across annotation rounds separated by weeks or months. Section~\ref{sec:consistency} reports empirical checks on how well that objective was met.

  Annotation conformed to the UD guidelines \autocite{nivreUniversalDependenciesV12016}, and the transcription conventions follow the project's knowledge base.\footnote{\url{https://kb.dalme.org/en/user_guide/transcription}} The following are project-specific decisions that the UD specification does not uniquely determine and that recur in this material. They are documented here because they are the loci where a second annotator would be most likely to diverge, and because Section~\ref{sec:residual-errors} shows that roughly one third of the final model's residual errors fall on these distinctions.

  \paragraph{Surface forms are preserved, lemmata are regularized.} No orthographic normalization is applied to token surface forms, since the variation described in Section~\ref{sec:challenges} is treated as evidence rather than noise. Lemmata are regularized to a canonical form so that orthographic variants can be counted together, thus \emph{condam} and \emph{quondam} share a lemma while both spellings remain in the transcription \autocite{bollmannLargescaleComparisonHistorical2019}.

  \paragraph{Vernacular items inherit Latin morphosyntax.} Occitan vernacular lexemes that appear inflected in Latin forms and that function within Latin clauses are annotated according to the Latin morphosyntactic requirements of their contexts. The vernacular status of the lexeme is recorded in the lexicon rather than in the syntactic tree. This design preserves internal consistency of the syntactic layer at the cost of making code-switching invisible in the CoNLL-U output. The lexicon's language labels (Section~\ref{sec:lexicon-rebuild}) compensate for this trade-off \autocite{cetinogluChallengesComputationalProcessing2016}.

  \paragraph{Material genitives attach as nominal modifiers while material adjectives are adjectival modifiers.} Trailing genitives that denote material or provenance attach as nominal modifiers to the object they qualify (\emph{e.g.} \emph{item unum morterium lapidis}). When equivalent information is expressed adjectivally it is treated as an adjectival modifier, even though Latin permits substantive readings and inventory usage encourages them (\emph{e.g.} \emph{scutellis terreis}). These distinctions are not always determinate, and both directions of disagreement persist in the final model's errors.

  \paragraph{Multi-word toponyms and personal names are analyzed as flat sequences.} Sequences such as \emph{carreria Sancti Augustini} are annotated as flat name structures rather than head-plus-modifier, except where overt inflectional agreement indicates internal syntactic structure.

  \paragraph{Verbless \emph{item} clauses are headed by their object.} The enumerative frame of the DALME-Marseille corpus leaves the majority of clauses without an overt predicate. In such cases the enumerated object is taken as the clause head, \emph{item} attaches as a discourse marker, and quantifiers attach to the object rather than to each other (no null predicate is posited).

  Lastly, annotation was limited to the basic dependency layer and no enhanced dependency annotations were produced (see Section~\ref{sec:syntactic-metrics}). Thus, these conventions apply to the sole syntactic layer represented in the gold standard and model outputs.

  \subsection{Annotation Consistency}
  \label{sec:consistency}

  All annotation was performed by a single annotator—the author—who has expertise in both Latin and dependency syntax. This choice determines the scope of available quality controls since inter-annotator agreement cannot be measured \autocite[556]{artsteinSurveyArticleIntercoder2008}.

  The formulaic nature of the corpus means that identical word sequences recur across rounds separated by weeks or months, and the annotator had no record of prior decisions when later instances were corrected, thus making \emph{intra-annotator drift} the principal residual risk. The appropriate control for this is therefore \emph{intra-annotator consistency} over time.

  \begin{table*}[t]
 \centering
 \sffamily\footnotesize
 \begin{tabularx}{\textwidth}{Xrrrrrr}
  \toprule
  \scriptsize\textbf{Criterion} & \scriptsize\textbf{Groups} & \scriptsize\textbf{Decisions} & \scriptsize\textbf{~~~UPoS~~~} & \scriptsize\textbf{~~~Head~~~} & \scriptsize\textbf{Deprel} & \scriptsize\textbf{~~~LAS~~~} \\
  \midrule
  Exact repeats & 64 & 286 & 99.30 & 100.00 & 99.65 & 99.65 \\
  $\le$1 substitution, $\ge$70\% similar & 116 & 615 & 99.51 & 99.67 & 99.67 & 99.51 \\
  $\le$2 substitutions, $\ge$70\% similar & 120 & 635 & 99.53 & 99.69 & 99.69 & 99.53 \\
  \midrule
  $\le$2 substitutions, $\ge$50\% similar & 105 & 1,877 & 99.79 & 87.80 & 99.57 & 87.59 \\
  \bottomrule
  \end{tabularx}
  \caption{Intra-annotator consistency, as percentage agreement between analyzes assigned to the same or nearly the same word sequence in different annotation rounds. The last row is a boundary check rather than a result since at 50\% similarity the groups are no longer duplicates, and head agreement collapses accordingly.}
  \label{tab:consistency}  
\end{table*}

  Across the 1,804 corrected sentences and the 200-sentence gold standard, 64 distinct word sequences recurred in different rounds. Comparing the analyzes assigned to these recurrences yields 286 paired token decisions. Agreement on these paired decisions was 99.30\% for UPoS, 100\% for head attachment, and 99.65\% for dependency relation (and for the two combined).\footnote{Cohen's $\kappa=0.990$ over eight categories \autocite[40]{cohenCoefficientAgreementNominal1960}} Only two sequences were analyzed differently on different occasions, and both divergences were analytic choices rather than errors (\emph{quoquina}, once tagged as a noun and once as an adverb, and \emph{lapidis}, once adjectival and once nominal).

  Exact repetition is a favourable test case and therefore a limited one. Expanding the comparison to \emph{near}-duplicates (sequences differing by one or two substitutions) approximately doubles the sample while preserving high agreement (Table~\ref{tab:consistency}).\footnote{Two methodological choices merit brief comment. First, the similarity threshold is relative, rather than an absolute edit budget. At two substitutions, superficially dissimilar pairs can qualify as near-duplicates if they share high-frequency tokens such as \emph{item}, which confounds interpretation. Second, counting all pairwise combinations induces a quadratic, non-independent set of comparisons and artificially inflates apparent sample size, thus the analysis counts annotation events pairwise to avoid this artefact. When the similarity threshold is relaxed too far, head-agreement falls (to 87.8\%), showing the measure's sensitivity to the matching criterion.}

  These results measure consistency, not inter-annotator agreement, thus they establish that a single annotator applied the scheme stably over time, not that a second annotator would have applied it identically. In addition, repetition is precisely where stability is easiest, so these figures are an upper bound on corpus-wide consistency. They nonetheless provide a meaningful control given the project's constraints.

  Two additional structural controls further reduced sources of inconsistency. First, the sentence batches were disjoint, so no sentence was annotated twice within the training data. Second, the annotation followed the UD inventory, rather than a bespoke tagset, thus constraining the set of available analytical choices and limiting arbitrary variation (a particularly valuable constraint for the elliptical constructions prevalent in this corpus).

  \subsection{Training and Data Partitioning}
  \label{sec:partitioning}

  All models were trained with the Stanza framework using its standard UD training scripts without architectural modification \autocite{qiStanzaPythonNatural2020}.\footnote{The models were trained on a single AWS \token{g4dn.xlarge} instance (4 vCPUs on 2 physical cores at 2.5\,GHz, 16\,GB of memory, one NVIDIA T4 GPU with 16\,GB) running the Deep Learning Base AMI for Amazon Linux 2, version 41.0. The requirements are modest by current standards (see Section~\ref{sec:limitations}).} The experiment therefore inherited Stanza's default architectures, namely a bidirectional LSTM tagger for the universal and language-specific tag sets and the morphological features, a sequence-to-sequence lemmatizer fronted by the domain dictionary described in Section~\ref{sec:lexicon-rebuild}, and a deep biaffine graph-based dependency parser \autocite{dozatDeepBiaffineAttention2017}.

  Hyperparameters were left at Stanza's defaults except in three respects. The parser used a batch size of 2,500 rather than 5,000, the parser's early-stopping patience was 5,000 steps rather than 3,000, and the lemmatizer was trained for 120 rather than 60 epochs. The random seed was Stanza's default (1234) throughout. No hyperparameter was tuned to optimize performance on this corpus, and none varies between iterations. That invariance is intentional: by holding the training configuration fixed across iterations, observed differences amongst models are attributable to the data rather than to configuration drift.

  Pre-trained word vectors were 300-dimensional embeddings trained on the corpus itself and were rebuilt twice during the experiment. One set served iterations $s_1$–$s_5$, a second $s_6$–$s_7$, and a third $s_8$–$s_9$. All three cover essentially the same vocabulary (2,855 of roughly 2,860 types are shared), but they are independent training runs rather than incremental refinements. Stanza treats these vectors as fixed input features during model training while learning a separate 75-dimensional task embedding, so the vectors enter the models as external features rather than as trainable parameters. Section~\ref{sec:adp-dip} returns to the rebuild points, which coincide with two prominent steps in the learning curve.

  Training was data-cumulative and always begun from scratch. At iteration $n$, the corrected sentences from batches $s_1$ through $s_n$ were pooled, shuffled, and split into training, development, and held-out portions as specified below, and a fresh set of models was trained on the training portion. Models were \emph{not} fine-tuned from their predecessors, with each iteration beginning from the same starting point and differing only in the amount of in-domain data it received.

  This design conflates two factors, since each successive model is trained on more data \emph{and} is retrained from scratch, so the trajectory reported in Section~\ref{sec:iteration-results} measures the combined effect. An incremental fine-tuning design would isolate per-batch contributions more precisely but would also accumulate any drift introduced by a sequence of updates. Retraining was chosen for reproducibility and, consequently, the reported curve is best read as a curve of dataset size rather than of incremental adaptation.

  For iteration $n$, the pool of corrected batches $s_1\ldots s_n$ was shuffled and split 90/10 into a training portion and a held-out test portion. The pool grows from 204 sentences at $s_1$ to 1,804 at $s_9$. Stanza's data-preparation step expects an explicit development set and, if none is provided, it splits the 90\% portion again (80/20 with a fixed seed) to create one. Thus, the effective training set is 72\% of the pooled sentences rather than 90\%. At $s_1$, this yields 148 training sentences, 36 development sentences, and 20 held-out sentences, while at $s_9$ the corresponding figures are 1,300, 324 and 180. Where cumulative sentence counts are reported, they refer to the pooled quantity on which annotation effort was expended, of which the model saw approximately seven-tenths.

  No regularization beyond Stanza's defaults was applied, and no cross-validation was performed. With a single held-out set of 200 sentences and nine models to compare, $k$-fold procedures would have added little to the direct comparison, and the fixed gold standard described in Section~\ref{sec:gold} has the advantage that all nine iterations and all five baselines are evaluated on identical material.

  \subsection{Evaluation Design}

  Evaluation was conducted against a fixed gold standard and utilized metrics in three categories: morphological accuracy, syntactic performance, and annotation effort.

  \subsubsection{The Gold Standard}
  \label{sec:gold}

  The evaluation set comprises 200 sentences (1,441 tokens and 1,241 non-punctuation words) drawn from the corpus and annotated independently of the training batches. Construction proceeded as follows: surface forms, lemmata, and UPoS tags were exported from the database; the text was parsed to provide an initial syntactic layer; the output was converted to \emph{brat} standoff format at ten sentences per document and corrected by hand; and the corrected files were exported to CoNLL-U for evaluation.

  \subsubsection{Morphological Metrics}

  Morphological evaluation encompassed part-of-speech tagging, prediction of morphological features, and lemmatization. The measures employed are those defined in Section~\ref{sec:metrics}, applied as follows.

  Tagging and feature accuracy were computed at token level against the fixed gold standard. Section~\ref{sec:adp-dip} additionally reports accuracy broken down by gold PoS category, which permits interpretation of the chain's early behaviour when aggregate measures concealed a substantial collapse in one category accompanied by immediate gains in the two dominant ones.

  Lemmatization was scored by exact string match against the gold-standard lemma. Lemmatizer scores are reported separately from other morphological tasks because lemmata are supplied predominantly by the domain lexicon (Section~\ref{sec:lexicon-rebuild}) and conflating these mechanisms would overstate the loop's contribution. Section~\ref{sec:oov} further disaggregates lemmatization by vocabulary status and by language \autocite{muellerEfficientHigherorderCRFs2013,kestemontLemmatizationVariationrichLanguages2017}.

  The out-of-vocabulary rate was computed with respect to the set of distinct lemmata in the gold standard, measured against each model's training vocabulary. It is reported alongside accuracy because it is determined solely by the training data and is therefore independent of the lexicon lookup and regularization procedures described above.

  \subsubsection{Syntactic Metrics}
  \label{sec:syntactic-metrics}

  The syntactic measures defined in Section~\ref{sec:metrics} were applied following the CoNLL 2018 conventions to ensure that results are comparable with published UD evaluations \autocite{zemanCoNLL2018Shared2018}. Two implementation points merit emphasis.

  No enhanced dependency scores are reported. Enhanced Universal Dependencies—in the sense of \textcite[2372]{schusterEnhancedEnglishUniversal2016}—require an annotation layer distinct from the basic tree, adding and augmenting relations to make implicit connections between content-words explicit. Neither the gold standard nor the model outputs include such an enhanced layer (the DEPS field in both simply restates the basic head and label), so any \enquote{enhanced} LAS computed would be identical to LAS and therefore vacuous. Constructing an enhanced annotation layer for this corpus would be valuable—particularly for the elliptical and coordinate phenomena characteristic of inventory syntax—but it is beyond the present scope and treated as future work.

  To complement aggregate metrics, syntactic performance by construction type is also reported. The 200-sentence gold standard was partitioned—using surface heuristics from Section~\ref{sec:corpus-comparison}—into coarse classes (simple versus complex object descriptions, single- versus multiple-object listings, and code-switching). Per-class scores test whether the method merely captures the formulaic core or also improves performance on structurally and lexically harder material. Some classes are small and therefore yield noisy estimates, a point that is noted where relevant.

  \begin{table*}[t]
 \centering
 \sffamily\footnotesize
 \begin{tabularx}{\textwidth}{XXXccccc}
  \toprule
  \scriptsize\textbf{Iteration} & \scriptsize\textbf{Sentences} & \scriptsize\textbf{Hours} & \scriptsize\textbf{~~~~UPoS~~~~} & \scriptsize\textbf{~~~~XPoS~~~~} & \scriptsize\textbf{Features} & \scriptsize\textbf{Lemmata} & \scriptsize\textbf{All Tags} \\
  \midrule
  Baseline (ITTB) & — & 0.0 & .801 & .660 & .640 & .587 & .570 \\
  \normalsize{$s_{1}$} & 204 & 8.8 & .743 & .503 & .568 & .948 & .472 \\
  \normalsize{$s_{2}$} & 404 & 14.1 & .718 & .527 & .587 & .943 & .502 \\
  \normalsize{$s_{3}$} & 604 & 18.3 & .753 & .614 & .709 & .945 & .589 \\
  \normalsize{$s_{4}$} & 804 & 21.5 & .817 & .693 & .743 & .947 & .670 \\
  \normalsize{$s_{5}$} & 1,004 & 24.3 & .829 & .689 & .740 & .958 & .676 \\
  \normalsize{$s_{6}$} & 1,204 & 26.9 & .917 & .842 & .854 & .954 & .831 \\
  \normalsize{$s_{7}$} & 1,404 & 29.4 & .922 & .849 & .865 & .956 & .833 \\
  \normalsize{$s_{8}$} & 1,604 & 31.6 & .977 & .913 & .920 & .958 & .904 \\
  \normalsize{$s_{9}$} & 1,804 & 33.2 & .977 & .920 & .926 & .953 & .915 \\
  \bottomrule
  \end{tabularx}
  \caption{Morphological performance across the nine reported iterations, F1 on the 200-sentence gold standard. Sentences are the pooled corrected seed data the model was trained on, hours are cumulative annotation time. The ITTB baseline is trained on 26,977 sentences of out-of-domain text and required no annotation of this corpus.}
  \label{tab:morph-curve}  
\end{table*}

  \subsubsection{Annotation Effort}

  Annotation effort was quantified using two complementary, independent measures. The primary measure is the elapsed time recorded for each round, which is a single aggregate duration covering the entire correction pass for the sampled batch. This coarse-grained recording yields a per-round rate (minutes per sentence) and an overall cumulative time. It does not disaggregate morphological and syntactic corrections, distinguish amongst construction types, nor account for ancillary activities such as tooling, data conversion, or model training.

  The secondary measure is timing-independent and therefore recoverable retrospectively: the proportion of pre-annotated tokens whose head assignment or dependency label differs between the pre-annotation and the corrected output. This token-level correction rate is resilient to biases in self-reported timing and, on this corpus, correlated closely with the elapsed-time measurements. Both measures are reported in Section~\ref{sec:iteration-results} because they are methodologically independent and permit mutual validation.

  \subsubsection{Model Validation}

  Each model was evaluated against the fixed 200-sentence gold standard described in Section~\ref{sec:gold}. The statistical significance of the improvements between iterations was assessed using bootstrap resampling over sentence-level scores ($n=10{,}000$), complemented by paired $t$-tests as a parametric check. Both procedures are reported to mitigate the risk of spurious confidence associated with a modest evaluation set and to allow cross-validation when the two methods concur \autocite{drorHitchhikersGuideTesting2018}.

  \subsection{Stopping}
  \label{sec:stopping}

  No formal convergence criterion governed the experiment. Annotation stopped after nine rounds, after the annotator judged subjectively that the pre-annotation had become sufficiently accurate that correction was no longer teaching the model substantially. That judgement was retroactively borne out by the correction rate reaching a plateau after the fourth round (Table~\ref{tab:annotation-effort-efficiency}). The analysis of stopping points in Section~\ref{sec:iteration-results} is therefore \emph{post hoc}. It describes where the returns in fact diminished, which is useful for planning comparable projects, but it did not steer this one.

  Prospectively, two inexpensive signals tracked the expensive one closely. The correction rate—the proportion of pre-annotated tokens the annotator changed—is computable at the end of each round, requires no held-out gold standard, and flattened at the same point that the evaluation metrics began to plateau. Annotation time per sentence behaved similarly. A project without the budget to build an independent gold standard before it starts could reasonably use either as a stopping signal, and they are recommended on the strength of their behaviour here.

  \section{Results}
  \label{sec:results}

  This section reports what the iterative procedure achieved on the DALME-Marseille corpus, including the trajectory of nine successively trained models and their comparison with the baselines of Section~\ref{sec:baselines}.

  \begin{table*}[t]
 \centering
 \sffamily\footnotesize
 \begin{tabularx}{\textwidth}{XXXccccc}
  \toprule
  \scriptsize\textbf{Iteration} & \scriptsize\textbf{Sentences} & \scriptsize\textbf{Hours} & \scriptsize\textbf{~~~~UAS~~~~} & \scriptsize\textbf{~~~~LAS~~~~} & \scriptsize\textbf{~~~~CLAS~~~~} & \scriptsize\textbf{~~~~MLAS~~~~} & \scriptsize\textbf{~~~~BLEX~~~~} \\
  \midrule
  Baseline (ITTB) & — & 0.0 & .648 & .482 & .425 & .168 & .258 \\
  \normalsize{$s_{1}$} & 204 & 8.8 & .767 & .638 & .643 & .342 & .606 \\
  \normalsize{$s_{2}$} & 404 & 14.1 & .783 & .617 & .623 & .377 & .589 \\
  \normalsize{$s_{3}$} & 604 & 18.3 & .817 & .647 & .644 & .457 & .609 \\
  \normalsize{$s_{4}$} & 804 & 21.5 & .815 & .704 & .692 & .521 & .657 \\
  \normalsize{$s_{5}$} & 1,004 & 24.3 & .845 & .719 & .704 & .544 & .673 \\
  \normalsize{$s_{6}$} & 1,204 & 26.9 & .861 & .813 & .800 & .686 & .759 \\
  \normalsize{$s_{7}$} & 1,404 & 29.4 & .861 & .817 & .793 & .684 & .747 \\
  \normalsize{$s_{8}$} & 1,604 & 31.6 & .955 & .924 & .904 & .811 & .859 \\
  \normalsize{$s_{9}$} & 1,804 & 33.2 & .945 & .917 & .897 & .822 & .845 \\
  \bottomrule
  \end{tabularx}
  \caption{Syntactic performance across the nine reported iterations, F1 on the 200-sentence gold standard. Sentences and hours as in Table~\ref{tab:morph-curve}.}
  \label{tab:syn-curve}  
\end{table*}

  \subsection{Performance Across Iterations}
  \label{sec:iteration-results}

  \subsubsection{Morphological Performance}

\begin{figure}[H]
    \centering
    \includegraphics[alt={Line chart of morphological and syntactic accuracy across the nine reported iterations against the ITTB baseline, with a shaded band marking the early rounds whose tagging accuracy falls below that baseline.}, width=\columnwidth,height=20\gridunit,keepaspectratio]{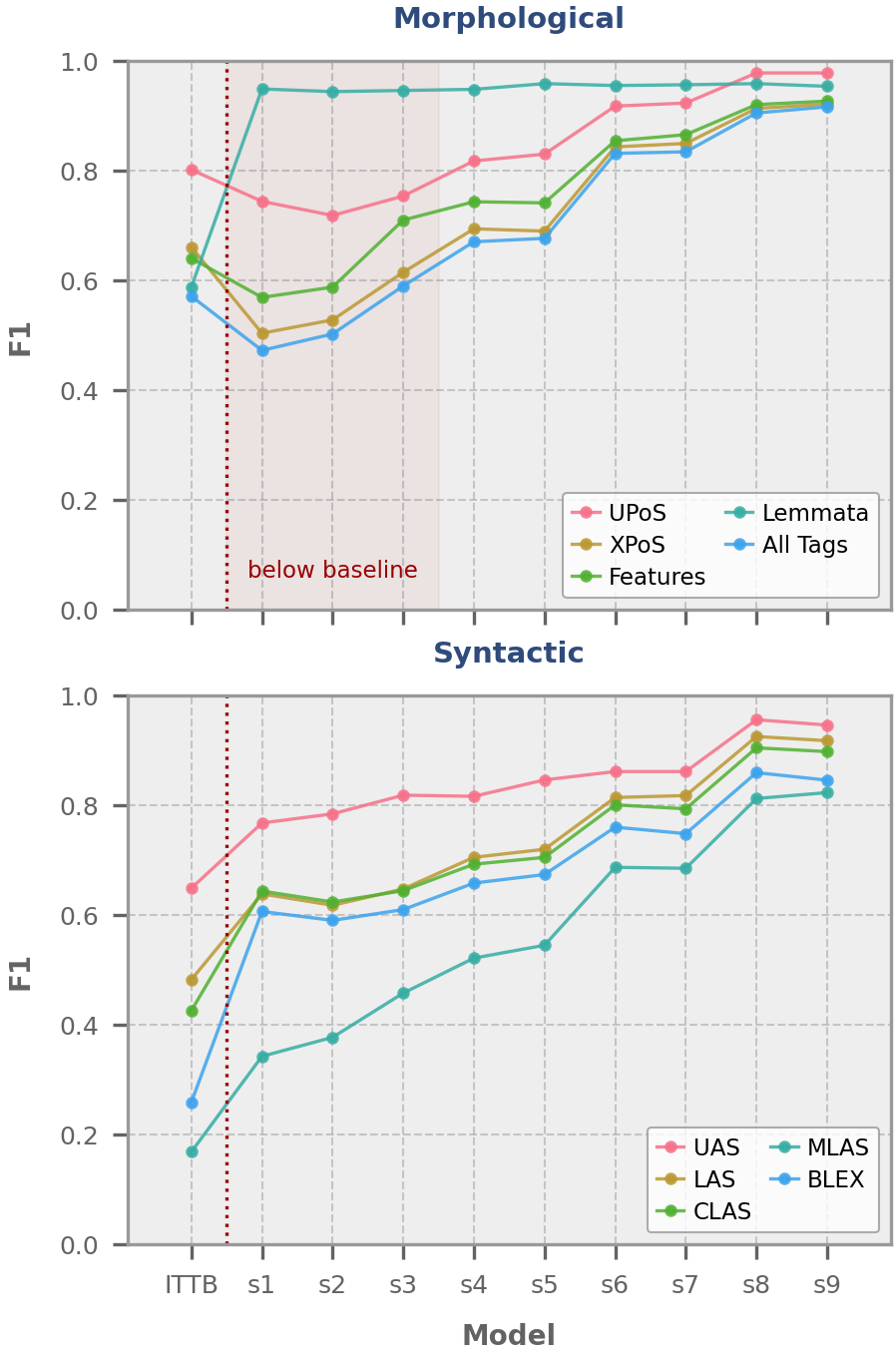}
    \caption{Performance across the nine reported iterations against the ITTB baseline. The shaded band marks the rounds whose UPoS accuracy falls below the out-of-domain baseline, analyzed in Section~\ref{sec:adp-dip}. Lemmata is flat throughout, for the reason given in Section~\ref{sec:results}.}
    \label{fig:learning-curve}  
  \end{figure}

  Table~\ref{tab:morph-curve} summarizes the morphological trajectory, and Figure~\ref{fig:learning-curve} plots it alongside the syntactic curve. Three observations are salient.

  \paragraph{Substantial tagging gains, but non-monotonic progression.} UPoS accuracy increased from 0.743 at $s_1$ to 0.977 at $s_9$. The language-specific tag set, the morphological features, and the combined All Tags measure likewise rose markedly, by between 63\% and 94\% (0.503 to 0.920, 0.568 to 0.926, 0.472 to 0.915 respectively). The increase of All Tags from 0.570 (baseline) to 0.915 at $s_9$ is the clearest aggregate indicator of the procedure's effectiveness.

  The improvement was not smooth. The first three adapted models performed below the ITTB baseline on UPoS (0.743, 0.718, 0.753 versus 0.801) and only exceeded it by $s_4$. Section~\ref{sec:adp-dip} attributes this early dip chiefly to one PoS category. Two pronounced step changes occurred later (between $s_5$ and $s_6$, and between $s_7$ and $s_8$), so the learning curve is better characterized as plateaus separated by discrete gains than as steady convergence.

  \paragraph{Lemmatization remains stable.} The Lemmata score was essentially flat (0.948 at $s_1$ to 0.953 at $s_9$, with minor fluctuations in between). Bootstrap resampling over sentence-level scores fails to reject the null of no change (\emph{e.g.} $p=0.88$ for $s_9$ versus $s_1$), and paired $t$-tests concur. This stability reflects the dominant role of the domain lexicon at initialization, since lemmata are already well covered at $s_1$, leaving little scope for iterative training to improve exact-match lemmatization. Section~\ref{sec:oov} examines the residual errors.

  \subsubsection{Syntactic Performance}

  Syntactic performance improved steadily and did not fall below the baselines (Table~\ref{tab:syn-curve}). UAS increased from 0.648 to 0.945 and LAS from 0.482 to 0.917, while MLAS exhibited the steepest relative gain (0.244 to 0.822, a 237\% relative improvement). MLAS is the metric on which the adapted models most clearly outperformed models trained on other treebanks.

  All of these improvements are statistically significant by bootstrap resampling over sentence-level scores ($n=10{,}000$) and by paired $t$-tests (\emph{e.g.} $p<0.001$ for $s_9$ versus $s_1$ across the principal measures). Only the Lemmata column does not reach significance in these comparisons.

  Two minor irregularities are notable, with LAS falling slightly between $s_1$ and $s_2$ (0.638 to 0.617) before resuming its ascent, and the final round exhibiting a marginal decline in UAS and LAS (0.955 to 0.945 in UAS, 0.924 to 0.917 in LAS) while MLAS continues to improve. On a 1,441-token gold standard, such swings correspond to roughly ten tokens and lie within expected sampling variance.

  \begin{table*}[t]
 \centering
 \sffamily\footnotesize
 \setlength{\tabcolsep}{5pt}
 \begin{tabularx}{\textwidth}{X|rr|cccccc|ccccc}
  \toprule
   & \multicolumn{2}{c|}{\scriptsize\textbf{Training}} & \multicolumn{6}{c|}{\scriptsize\textbf{Morphological}} & \multicolumn{5}{c}{\scriptsize\textbf{Syntactic}} \\
  \cmidrule{2-14}
  \scriptsize\textbf{Model} & \scriptsize\textbf{Sent.} & \scriptsize\textbf{Tokens} & \scriptsize\textbf{UPoS} & \scriptsize\textbf{XPoS} & \scriptsize\textbf{Feats.} & \scriptsize\textbf{Lem.} & \scriptsize\textbf{Ens.} & \scriptsize\textbf{All} & \scriptsize\textbf{UAS} & \scriptsize\textbf{LAS} & \scriptsize\textbf{CLAS} & \scriptsize\textbf{MLAS} & \scriptsize\textbf{BLEX} \\
  \midrule
  ITTB & 26,977 & 450,554 & .801 & .660 & .640 & .587 & .782 & .570 & .648 & .482 & .425 & .168 & .258 \\
  LLCT & 9,023 & 242,431 & .850 & .550 & .528 & .554 & .794 & .487 & .731 & .622 & .586 & .223 & .264 \\
  Perseus & 2,273 & 29,574 & .756 & .514 & .574 & .757 & .925 & .471 & .577 & .407 & .391 & .151 & .287 \\
  PROIEL & 18,689 & 205,566 & .727 & .531 & .662 & .656 & .790 & .493 & .425 & .285 & .267 & .150 & .219 \\
  UDante & 1,723 & 55,818 & .849 & .643 & .550 & .711 & .915 & .522 & .697 & .566 & .523 & .244 & .370 \\
  \midrule Best of 5 & — & — & .850 & .660 & .662 & .757 & .925 & .570 & .731 & .622 & .586 & .244 & .370 \\
  \rowcolor{black!8} \textbf{\normalsize{$s_{9}$}} & 1,804 & 14,516 & .977 & .920 & .926 & — & .953 & .915 & .945 & .917 & .897 & .822 & .845 \\
  \textbf{$\Delta$} & — & — & +.127 & +.260 & +.264 & — & +.028 & +.345 & +.214 & +.294 & +.310 & +.578 & +.474 \\
  \bottomrule
  \end{tabularx}
  \caption{The final reported model against every treebank baseline on the DALME-Marseille gold standard, over the same metrics as Table~\ref{tab:pre-trained-performance}. Every column uses the plain models except \emph{Ens.}, which uses the ensemble of the sequence-to-sequence lemmatizer with domain lexicon lookup for every row alike. \normalsize\normalsize{$s_{9}$}\scriptsize~is only ever run in that configuration, so it has no plain lemmatization figure to report. \emph{Best of 5} is the highest of the five treebank models in each column, which is not the same model in every column, and $\Delta$ is the final model's margin over it. Training sentences and tokens for~\normalsize\normalsize{$s_{9}$}\scriptsize~are the pooled corrected sentences, for the baselines, the size of the treebank the model was trained on is used.}
  \label{tab:cross-treebank-performance}  
\end{table*}

  \subsection{Cross-Treebank Performance Comparison}

  This comparison is meant to address the question of whether any existing Latin model, given sufficient adaptation, could have achieved comparable performance (Table~\ref{tab:cross-treebank-performance}).

  The final trained model outperforms every baseline on every reported metric, despite being trained on 14,516 in-domain tokens versus ITTB's 450,554 tokens—approximately 97\% less data. The gains are largest where the baselines were weakest, with All Tags increasing by 34.5 points over the best baseline (0.570 to 0.915), MLAS by 57.8 points (0.244 to 0.822, a 237\% relative gain), and LAS by 29.4 points.

  The sole, narrow exception is lemmatization, where the best baseline (Perseus with the ensemble) attains 0.925 versus the final model's 0.953, a gap of 2.8 points. This result is consistent with the analysis above. The same domain lexicon underlies the lemmatization column for all configurations, thus that column primarily reflects lexicon coverage rather than differences in model learning.

  A practical implication that follows directly from the comparison is that in-domain annotated data is vastly more valuable per token than out-of-domain data. Roughly 1,804 sentences of corrected inventory text produce a better parser for this corpus than 26,977 sentences of Aquinas. The direction of this conclusion is expected, but the magnitude is striking and supports the method's cost-effectiveness.

  \begin{figure*}[t]
    \centering
    \includegraphics[alt={Line chart of tagging accuracy by gold-standard category across iterations. Adpositions collapse from 98.1 per cent at the baseline to 3.7 per cent at iteration 2 before recovering to 97.2 per cent by iteration 6.}, width=\textwidth]{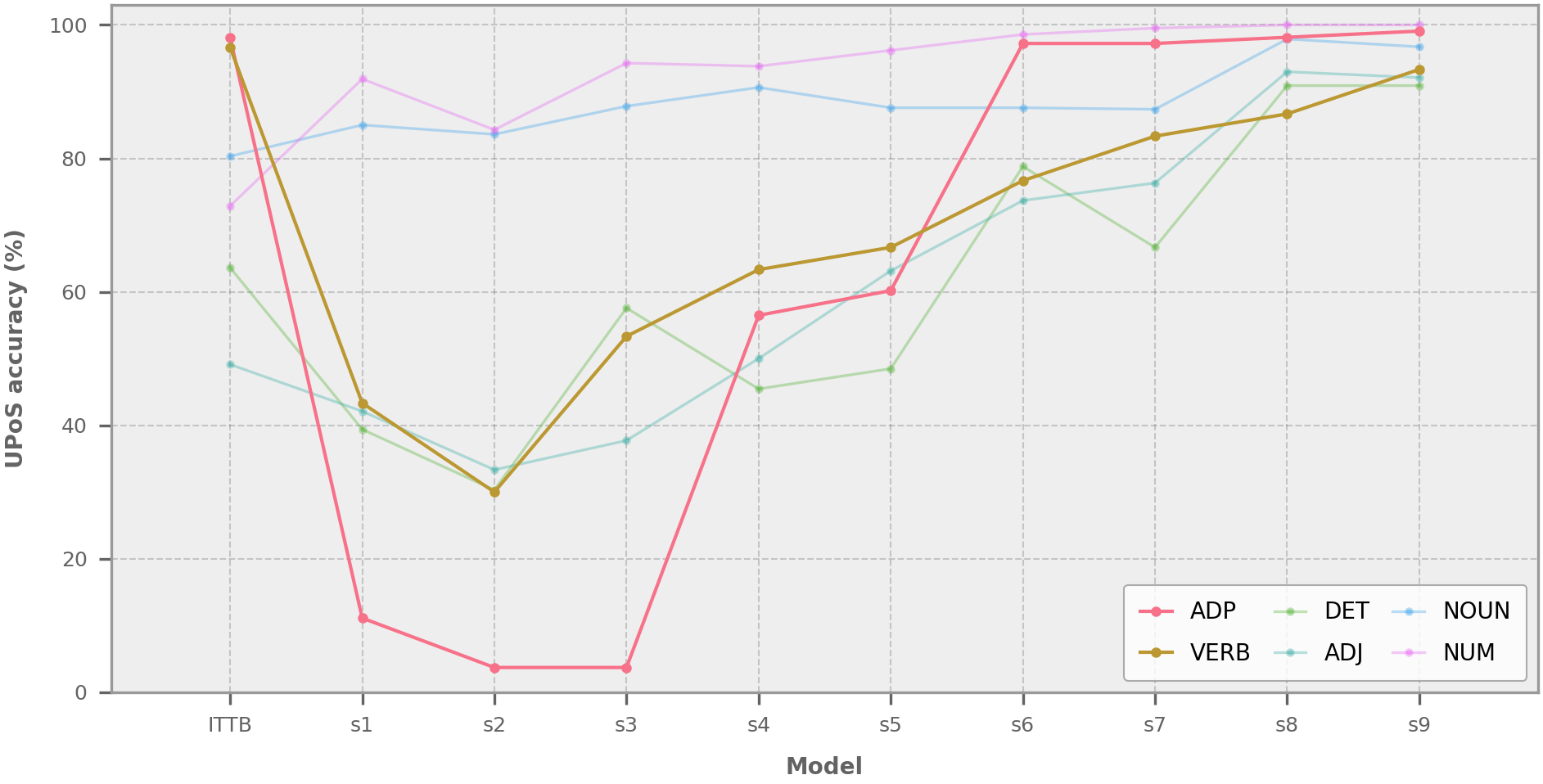}
    \caption{Tagging accuracy by gold-standard category. Adpositions fall from 98.1\% at the baseline to 3.7\% at \normalsize{$s_2$}\scriptsize~before recovering to 97.2\% at \normalsize{$s_6$}\scriptsize, verbs follow the same shape over far fewer tokens. The two categories that dominate this corpus, nouns and numerals, improve from the first iteration instead. Categories with fewer than 30 gold tokens are omitted.}
    \label{fig:adposition-dip}  
  \end{figure*}

  \subsection{Learning Dynamics}

  \subsubsection{An Early Dip in Tagging}
  \label{sec:adp-dip}

  The most conspicuous feature of Table~\ref{tab:morph-curve}, and of the upper panel of Figure~\ref{fig:learning-curve}, is that the first three adapted models performed worse on UPoS tagging than the out-of-domain baseline used to initialize the loop. A per-category breakdown (Figure~\ref{fig:adposition-dip}) attributes the bulk of this effect to a single class. ITTB attains 98.1\% accuracy on adpositions, whereas the adapted models record 11.1\% at $s_1$, 3.7\% at $s_2$ and $s_3$, rising thereafter to 56.5\% at $s_4$, 60.2\% at $s_5$, and 97.2\% at $s_6$. Adpositions account for 108 of the 1,441 gold-standard tokens, so their swing explains approximately 7.1 of the 8.3 points lost between the baseline and $s_2$, \emph{i.e.} about 85\% of the observed dip. The principal mislabelling involves high-frequency prepositions reclassified as coordinating conjunctions or numerals (notably \emph{de} with 48 occurrences, \emph{in} with 23, \emph{cum} with 7, and, to a lesser extent, \emph{ad} and \emph{circa}).

  The remaining portion of the dip reflects the same redistribution phenomenon in smaller categories at $s_2$, where performance is lowest. For example, verbs decline from 96.7\% to 30.0\% (over 30 tokens), determiners from 63.6\% to 30.3\%, and adjectives from 49.1\% to 33.3\%.

  Concomitantly, the two categories that dominate the corpus improve immediately. At $s_1$, nouns rose from 80.3\% to 85.0\% and numerals from 72.9\% to 91.9\%. These gains partially offset the aggregate decline, which explains the apparently paradoxical arithmetic at $s_1$, where adpositions alone account for a 6.5-point decline while the net dip is 5.8 points because nominal categories move upward. In short, adaptation benefitted the nominal core from the outset, while the function-word periphery deteriorated temporarily.

  Two simple explanations are unlikely. Sparsity does not account for the effect, since adpositions constitute 8.99\% of the $s_1$ training tokens, marginally more than their 7.49\% share of the gold standard. Nor does annotation drift explain it: across all nine batches and the gold standard, tokens such as \emph{de} and \emph{in} are consistently annotated as adpositions and \emph{cum} predominantly as a subordinating conjunction, so the training labels are correct. The remaining explanation is model redistribution under a small, skewed training sample. A tagger exposed to roughly 1,280 in-domain words appears to reallocate its function-word probability mass in systematic ways that a larger sample later corrects.

  The two pronounced increases reflect the resolution of the same underlying failure mode. Between $s_5$ and $s_6$, the recovery was driven principally by adpositions (60.2\% to 97.2\%), with determiners and adjectives improving subsequently, while between $s_7$ and $s_8$, the improvement was broader and more content-focused, with nouns rising from 87.4\% to 97.9\%.

  One caveat is warranted, as it is not evident from the tabulated aggregates: the pre-trained word vectors were rebuilt between $s_5$ and $s_6$ and again between $s_7$ and $s_8$—exactly at the two boundaries where the curve exhibits step changes. Each rebuild coincides with the largest gain in the chain followed by an almost flat iteration. The vectors are unlikely to be the primary explanation. The per-category measurements that underpin the account above do not depend on inference from the aggregate curve, the three vector sets cover essentially the same vocabulary (so no lexical coverage was newly introduced), and the vectors are treated as fixed inputs rather than learned parameters. Nevertheless, the rebuilds are independent training runs and thus introduce an additional change beyond annotation quantity and, with only two such events, their contribution cannot be statistically disentangled. The overall trajectory remains robust—structural metrics improve across nine iterations and the endpoints are unaffected—but these two-step changes should not be attributed solely to accumulated annotation.

  This early dip illuminates the method's characteristic failure mode. The corpus's extreme nominal density, quantified in Section~\ref{sec:corpus-comparison}, predicts this asymmetry, with categories that are abundant in-domain adapting immediately, whereas categories that are relatively rare yet structurally important potentially deteriorating before recovering. Practitioners adapting models on a few hundred sentences of similarly skewed material should therefore expect a comparable curve and avoid interpreting an early drop as evidence of methodological failure.

  \subsubsection{Vocabulary Coverage}
  \label{sec:oov}

  The most direct indicator of domain adaptation in the training data is the out-of-vocabulary rate. With respect to the 316 distinct lemmata in the gold standard, the ITTB training vocabulary leaves 68.0\% out-of-vocabulary. After a single round of annotation, this proportion fell to 36.1\%, and by $s_9$ it reached 13.3\%—a reduction of 54.7 percentage points. This reduction was achieved by adding 1,158 lemma types to an initial vocabulary of 5,124. Because this measure is computed solely from the training data, it is not confounded by the lexicon issue discussed above.

  The improvements are heavily front-loaded, with $s_1$ alone accounting for more than half of the total reduction, while rounds $s_5$ through $s_9$ collectively move the rate only from 15.5\% to 13.3\%. This pattern provides the clearest empirical signal about where the marginal value of additional annotation lies in this setting.

  Improved lexical coverage does not, however, translate into commensurate gains in exact-match lemmatization. Accuracy on out-of-vocabulary lemmata increases from 49.0\% to 86.5\%, yet a persistent gap remains versus in-vocabulary items. At $s_9$, out-of-vocabulary accuracy was 86.5\% versus 97.3\% for in-vocabulary items, a penalty of 10.8 points (down only modestly from the baseline penalty of 17.9). The remaining out-of-vocabulary items are disproportionately vernacular, with Occitan accounting for 26.5\% of the baseline out-of-vocabulary lemmata, and 42.9\% of those remaining uncovered at $s_9$. In short, the vernacular component of the lexicon is the more recalcitrant problem, and the one which is the least assisted by further annotation of Latin material.

  \subsubsection{Code-Switching}

  The language composition of the gold standard mirrors that of the broader corpus (Section~\ref{sec:challenges}). At the sentence level, 73 of the 200 sentences are mixed-language. Mixed-language sentences are more difficult for every model, but the penalty declined with adaptation. The gap between Latin-only and code-switching sentences narrowed from 3.39 to 0.88 points on UPoS, from 7.93 to 3.91 on UAS, and from 6.34 to 3.64 on LAS. In other words, tagging accuracy nearly equalized the two classes, while parsing did not, and mixed sentences remained roughly four points harder at $s_9$.

  The lemmatization gap exhibits a different behaviour, at 1.46 points for the baseline and 1.49 points at $s_9$, it does not decrease. This pattern is expected if lemmata are supplied largely by a static lexicon rather than by a model that generalizes. On a per-token basis, Occitan tokens begin considerably behind Latin tokens for lemmatization (28.7\% versus 55.3\%) and end behind them (88.0\% versus 96.0\%), despite larger absolute improvements. These token-level figures are based on only 108 Occitan tokens in the gold standard and accordingly carry substantial uncertainty.

  \subsubsection{Construction Types}

  To identify which sentence types gained most from adaptation, the 200 gold-standard sentences were classified by construction type using surface heuristics based on the enumeration patterns described in Section~\ref{sec:corpus-comparison}, and accuracy per class was measured across the chain (Table~\ref{tab:construction-specific-improvements}).

  \begin{table}[H]
 \centering
 \sffamily\footnotesize
 \begin{tabularx}{3.04in}{Xrcccc}
  \toprule
   &  & \multicolumn{2}{c}{\scriptsize\textbf{UPoS}} & \multicolumn{2}{c}{\scriptsize\textbf{LAS}} \\
  \cmidrule(lr){3-4}\cmidrule(lr){5-6}
  \scriptsize\textbf{Const.} & \scriptsize\textbf{Sents} & \scriptsize\textbf{ITTB} & \scriptsize\textbf{\normalsize{$s_{9}$}} & \scriptsize\textbf{ITTB} & \scriptsize\textbf{\normalsize{$s_{9}$}} \\
  \midrule
  CSO & 80 & .845 & .992 & .521 & .989 \\
  CS & 73 & .783 & .973 & .438 & .889 \\
  CSO & 34 & .790 & .970 & .502 & .875 \\
  Other & 10 & .861 & .944 & .694 & 1.000 \\
  \midrule \multicolumn{6}{l}{\scriptsize\emph{Too few sentences to interpret:}} \\
  SMO & 2 & .882 & 1.000 & .765 & 1.000 \\
  CMO & 1 & .529 & 1.000 & .294 & .941 \\
  \bottomrule
  \end{tabularx}
  \caption{Accuracy by construction type on the gold standard, baseline versus final reported model. The last group is reported for completeness only. Legend: SSO = Simple, single object, CS = Code-switching, CSO = Complex, single object, SMO = Simple, multiple objects, CMO = Complex, multiple objects.}
  \label{tab:construction-specific-improvements}  
\end{table}

  Four classes contain sufficient examples to support comparative claims. Together, they account for 197 of the 200 sentences. Across these classes, those that begin the furthest behind tend to show the largest improvements. Code-switching sentences register a 19.0-point gain on UPoS tagging and a 45.1-point gain on LAS, while complex single-object descriptions, characterized by trailing relative clauses and prepositional qualifications, gain 18.0 and 37.4 points respectively. Simple enumerations, the largest class, both start and finish with the highest absolute scores (reaching 0.989 on LAS) and therefore have correspondingly less room for improvement.

  Two caveats are necessary. First, the improvements are measured against ITTB, which varies in strength across construction classes. Thus, the observed ordering partially reflects available headroom rather than purely differential learning. Second, two multi-object classes contain only one and two sentences respectively, so large percentage gains computed over such small denominators are not robust and should not be over interpreted.

  Despite these caveats, the results presented in Table~\ref{tab:construction-specific-improvements} support an important qualitative conclusion, namely that the method is not merely learning the formulaic core of the corpus. The class that improves most on both metrics is that defined by vernacular interference—the most challenging material in the corpus—and it approaches the simple enumerations on tagging within four points.

  \subsection{Cost and Benefit}
  \label{sec:cost-benefit}

  This section relates annotation effort to the accuracy gains observed.

  \begin{figure*}[t]
    \centering
    \includegraphics[alt={Scatter plot of accuracy against cumulative annotation hours, one point per reported iteration, showing most of the gain accruing in the first two thirds of the effort.}, width=\textwidth]{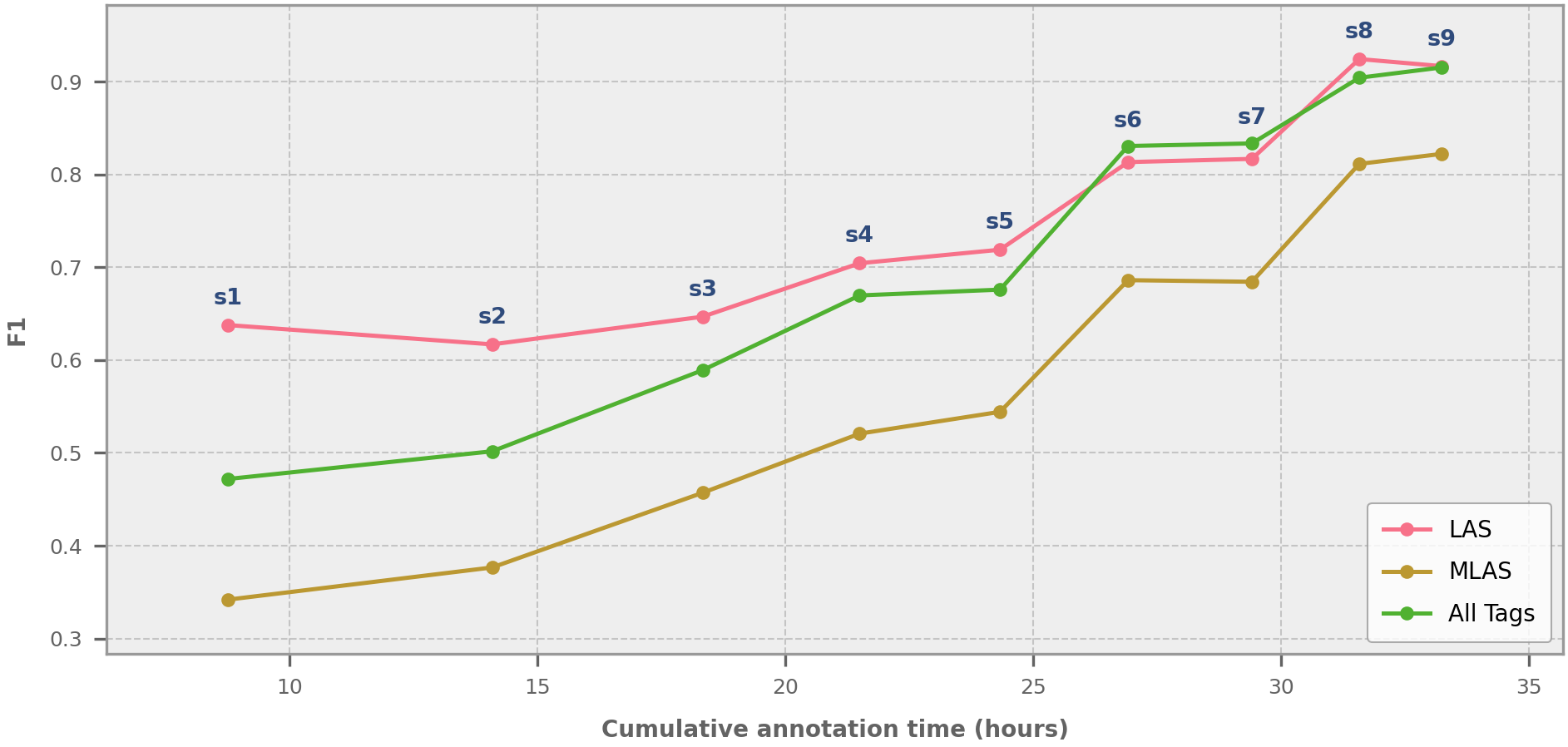}
    \caption{Accuracy against cumulative annotation effort. Points are the nine reported iterations, labelled by round. The horizontal axis is cumulative hours rather than iteration number.}
    \label{fig:cost-benefit}  
  \end{figure*}

  \subsubsection{Annotation Time}

  Annotation time per sentence declined substantially over the nine rounds, from 2.57 minutes at the first round to 0.50 minutes at the last (Table~\ref{tab:annotation-effort-efficiency}, the cumulative effort for the reported chain totals 33.2 hours).

  The token-level correction rate fell in parallel, from 54.3\% in the initial round to roughly 14–18\% from $s_5$ onwards, where it stabilized. The final five rounds fluctuated within that band rather than continuing to descend, with $s_9$ marginally above the floor. This non-zero plateau is informative, implying that some fraction of pre-annotation will always require manual adjustment because the source text genuinely underdetermines certain analyzes (Section~\ref{sec:residual-errors}). Practically, pre-annotation attained a useful standard after approximately four rounds, and subsequent rounds primarily yielded accuracy improvements rather than further increases in speed.

  \begin{table}[H]
 \centering
 \sffamily\footnotesize
 \begin{tabulary}{\textwidth}{CRRRRR}
  \toprule
  \scriptsize\textbf{Round} & \scriptsize\textbf{Sent.} & \scriptsize\textbf{Tokens} & \scriptsize\textbf{Mins.} & \scriptsize\textbf{Min/S} & \scriptsize\textbf{Corrected} \\
  \midrule
  \normalsize{$s_{1}$} & 204 & 1,979 & 525 & 2.57 & 54.3\% \\
  \normalsize{$s_{2}$} & 200 & 1,690 & 320 & 1.60 & 31.5\% \\
  \normalsize{$s_{3}$} & 200 & 1,566 & 255 & 1.28 & 23.2\% \\
  \normalsize{$s_{4}$} & 200 & 1,643 & 190 & 0.95 & 22.2\% \\
  \normalsize{$s_{5}$} & 200 & 1,480 & 170 & 0.85 & 14.1\% \\
  \normalsize{$s_{6}$} & 200 & 1,548 & 155 & 0.78 & 17.8\% \\
  \normalsize{$s_{7}$} & 200 & 1,496 & 150 & 0.75 & 15.6\% \\
  \normalsize{$s_{8}$} & 200 & 1,533 & 130 & 0.65 & 17.8\% \\
  \normalsize{$s_{9}$} & 200 & 1,581 & 100 & 0.50 & 18.0\% \\
  \midrule \textbf{Total} & 1,804 & 14,516 & 1995 & 1.11 & --- \\
  \bottomrule
  \end{tabulary}
  \caption{Annotation effort per round. \emph{Corrected} is the share of pre-annotated tokens whose head or relation the annotator changed, which is the quantity the bootstrapping loop is meant to drive down.}
  \label{tab:annotation-effort-efficiency}  
\end{table}

  \subsubsection{Returns to Annotation Effort}

  Relating cumulative annotation hours to LAS improvement (Figure~\ref{fig:cost-benefit}) shows uneven rather than continuous returns. Measured as LAS improvement per hour of annotation, the most productive rounds were $s_6$ (0.037 LAS points per hour) and $s_8$ (0.050 LAS points per hour). The rounds between and after these contributed comparatively little, and $s_9$ was marginally negative on this particular ratio while still producing gains in MLAS. Cumulatively, $s_8$ accounted for the entirety of the final LAS improvement and for 98\% of the MLAS improvement, consuming 31.6 of the 33.2 hours. Indeed, LAS was slightly higher at $s_8$ than at $s_9$ (0.924 versus 0.917), a difference equivalent to roughly ten tokens on the gold standard, and well within expected sampling variance (Section~\ref{sec:limitations}). Thus, $s_9$ should not be interpreted as a deterioration, but rather as having produced primarily morphological refinements instead of further structural gains.

  These figures should not be used to derive a general stopping rule. The signal is noisy at this scale—small token counts can shift metrics by non-trivial fractions of a point—and the two observed jumps correspond to the recovery of specific function-word categories (Section~\ref{sec:adp-dip}) rather than to a change in annotation practice. A narrower, more defensible conclusion is that, for a corpus as formulaic as the one studied here, most attainable gains accrued within roughly 1,200 annotated sentences and about 27 hours, while the final third of effort yielded refinement rather than transformation.

  While it can be directly stated that 33.2 hours of correction moved LAS from 0.482 to 0.917 on this corpus, comparing this chain to published totals for building a treebank from scratch is fraught, since those totals are often reported in heterogeneous units (person-months, tokens per hour, cost per sentence) and against annotation schemes of varying depth, so cross-project ratios would largely reflect choice of reference rather than substantive differences. What can be usefully transferred, provided the annotation scheme is specified, is the annotation rate. Averaged over the chain, the correction pass corresponds to 1.11 minutes per sentence (falling from 2.57 to 0.50) for a full morphosyntactic correction pass covering UPoS, XPoS, morphological features, lemma, head, and dependency relation. Projects adopting a shallower scheme should expect faster rates, while those annotating from scratch, rather than correcting machine output, should expect considerably slower ones.

  \subsection{Qualitative Analysis of Model Learning}
  \label{sec:residual-errors}

  The residual errors of the training are as revealing as its successes. Of the 1,441 tokens in the gold standard, 174 (12.1\%) carry at least one error in UPoS tag, lemma, head, or relation at $s_9$. If XPoS tags and morphological features are included, the count rises to 226 (15.7\%).\footnote{Errors overlap across categories—a token may be incorrect on more than one axis—so the percentages reported in this section exceed 100\% in sum.}

  \paragraph{Lemmatization is the largest residual error class.} Sixty-eight tokens (39.1\% of erroneous tokens) have an incorrect lemma. These errors are predominantly vernacular or technical vocabulary absent from the domain lexicon, for example \emph{sendar} for \emph{cendalum} (a silk fabric), or \emph{gallanda} for \emph{garlanda}. Tracking these forms through the chain demonstrates the loop's limited effect on them, with 51 of the 69 word forms lemmatized incorrectly at $s_1$ remaining incorrect at $s_9$.

  \paragraph{Syntactic errors are mostly genuine ambiguities.} One hundred and twenty tokens (69.0\% of erroneous tokens) have an incorrect head, an incorrect relation, or both. The most frequent confusions are nominal modifier versus adjectival modifier (8 cases), nominal modifier versus flat expression (7), adjectival modifier versus adnominal clause (7), and adnominal clause versus conjunct (6). These errors cluster in two recurring configurations. The first involves attachment within multi-word proper names. In \emph{carreria Sancti Augustini}, for example, the model interprets \emph{Sancti} as part of a flat name sequence, whereas the gold standard treats it as a nominal modifier—an analysis on which competent annotators may reasonably differ. The second configuration concerns clausal versus coordinate attachment in trailing qualifications that follow object descriptions, where medieval Latin frequently omits overt markers, leaving the correct analysis underdetermined by the surface text.

  \paragraph{Tagging errors are concentrated in one UPoS distinction.} Thirty-three tokens (19.0\%) carry an incorrect UPoS tag, and 19 of these involve the noun–adjective distinction. For instance, for \emph{scutellis terreis} the gold standard classifies \emph{terreis} as a noun while the model assigns it an adjectival tag. Latin routinely uses adjectives substantively, and the inventory genre—where objects are often identified by their material—renders the boundary between noun and adjective particularly fluid.

  Collectively, these patterns indicate that the model has learned the structural scaffolding of the documents and is parsing the enumerative frame reliably. Moreover, systematic baseline failures (\emph{e.g.} prepositions mislabelled as conjunctions or numerals) have been resolved. The remaining errors fall into two types: lexical items absent from the lexicon (which the loop touches only minimally) and genuine structural ambiguities the text does not disambiguate. Approximately one third of the residual errors are of a kind that a second annotator might plausibly resolve differently. At that point, further iteration is not the appropriate remedy and clearer annotation guidelines become the most effective resolution.

  \section{Discussion}

  Four questions follow from these results: whether the procedure generalizes to other materials; what the experiment implies for domain adaptation more broadly; for which tasks the reported scores suffice; and what conclusions the study cannot support.

  \subsection{Generalization to Other Corpora}
  \label{sec:generalizability}

  This study focused on a single-corpus and, therefore, its external generalizability is limited. It does, however, supply a reasonably precise account of why the procedure worked for this example, and the conditions for it working elsewhere can be read from that account.

  Three properties of this material were the most impactful:

  \begin{enumerate}
    \item It is formulaic, so a small annotated sample covers most of the structural variation present and the pre-annotation is already correct about most of a new batch before the annotator opens it.
    \item Its distance from existing resources is lexical and structural rather than distributional, so annotation supplies something that no quantity of additional out-of-domain Latin would have supplied.
    \item A usable starting model existed in ITTB, and although its All Tags score was low (0.570), correcting its output remained faster than annotating from scratch.
  \end{enumerate}

  The third condition is the one most easily overlooked, and it is a precondition rather than a convenience: a corpus meeting the first two, but not the third, cannot start the loop at all.

  \paragraph{Documentary administrative genres are the clearest candidates.} The most evident candidates are documentary administrative genres—tax records, court proceedings, commercial accounts, institutional registers—because their pragmatic function produces formulaic, recurrent structures, they use specialized object vocabularies absent from literary corpora \autocite[2]{korkiakangasLateLatinCharter2021}. Multilingual sources in which one language provides the grammatical frame and another supplies the lexical content (for example, trade records, diplomatic exchanges, and colonial administrative documents) exhibit the same configuration analyzed in Section~\ref{sec:challenges} and are therefore amenable to the same method. Morphologically rich historical varieties that possess an existing treebank but lack in-domain data meet the third precondition. The PROIEL family and related resources illustrate this case for a range of early Slavic and Germanic materials \autocite{eckhoffPROIELTreebankFamily2018}.

  \paragraph{Individually authored material fails the first condition.} Materials whose interest lies in their variety rather than their regularity, such as correspondence, literary prose, anything individually authored, would fail the first condition, and the correction rate is likely to plateau earlier and at a higher level. Languages with no treebank at all fail the third, as there is nothing to pre-annotate with, and a first batch would have to be built by hand before the loop could turn. Between these lies a wide middle ground where the question is empirical and straightforward to evaluate. A single batch of 200 sentences and one round of correction are sufficient to determine whether the pre-annotation is worth correcting. This requires a few working days, at most, rather than a substantial research programme \autocite{piotrowskiNaturalLanguageProcessing2012}.

  \paragraph{Practical limitations.} Computation is unlikely to constrain the procedure described in this study. The limiting factor, rather, is whether one person—or a group small enough to stay consistent—holds both the philological knowledge the material demands and enough syntactic training to apply a dependency scheme (Section~\ref{sec:limitations}). Where that combination must be assembled from several people, coordination costs are real and the consistency argument of Section~\ref{sec:consistency} no longer holds \autocite{fortAmazonMechanicalTurk2011}. Orthographic instability is a separate matter, and has a separate remedy, since it degrades lemmatization specifically, and a project facing it should budget for a lexicon rather than expect iteration to solve the problem \autocite{bollmannLargescaleComparisonHistorical2019}.

  \subsection{The Limits of Purely Computational Approaches}

  This paper evaluated how models, trained on the five available Latin treebanks, perform when applied directly to this corpus without further tuning. Those models perform poorly in this setting, and the best-performing baseline is not the one that an external typological reading of the corpus would have suggested. No alternative strategies, such as fine-tuning, model ensembling, or synthetic-data generation, were attempted, so the results should not be read as a negative assessment of those methods. Instead, what follows is a discussion of why such techniques would face intrinsic limitations on this target.

  The central obstacle is primarily lexical and structural rather than merely distributional. As Section~\ref{sec:genre-period} shows, transfer fails most instructively where it should have been feasible, since the corpus contains an object vocabulary, vernacular borrowings, and an elliptical enumerative frame that are largely absent from the available treebanks. Adaptation assumes the existence of a source domain from which the target is reachable, but methods that merely re-weight or redistribute existing knowledge are of limited utility when the target contains essentially novel lexical and constructional material \autocite{ben-davidTheoryLearningDifferent2010}.

  Data augmentation was not pursued for the same reason. Approaches such as synonym replacement, paraphrasing, or back-translation presuppose a generative model of the language capable of producing plausible historical variants, but for orthographically unstable medieval Latin no such reliable generator exists. Artificial augmentation risks introducing anachronistic or modern patterns rather than the historical variability the corpus exhibits \autocite{weiEDAEasyData2019,shortenSurveyImageData2019}.

  Only human expertise can supply the missing information, and the loop concentrates that expertise on the parts of the corpus that are the most informative. Correction by an expert supplies information that the models could not have obtained from more out-of-domain text, and the residual-error analysis in Section~\ref{sec:residual-errors} indicates fairly precisely what kind of information that is.

  Iterative correction demonstrably taught the models the recurrent syntactic constructions of the corpus—effectively a grammar of its formulaic frames—but it did not substitute for lexical coverage. Improving the lexicon requires external lexical resources and editorial effort, rather than further rounds of iterative correction \autocite{settlesActiveLearning2012}.

  The distinction merits emphasis because it specifies the expertise consumed by the loop. The corrections that materially improved model performance were syntactic judgements—\emph{e.g.} decisions about which of two plausible heads a trailing genitive takes, or whether a material term functions substantively—and these judgements generalize because the constructions recur. Lexical knowledge does not generalize in the same way. Knowing that \emph{gardacossium} denotes a garment or that \emph{sendar} denotes a silk fabric resolves those tokens alone. Both knowledge types are necessary for producing correct annotation, but only the syntactic judgements are primarily what the model learns.

  \subsection{Adequacy for Scholarly Use}

  Adequacy is use-dependent, and must be justified rather than asserted. A project presenting performance numbers should state explicitly the tasks for which those numbers suffice, since the same 0.92 LAS score that supports aggregate pattern-level queries does not justify claims about individual sentences \autocite{belinkovAnalysisMethodsNeural2019a,rogersPrimerBERTologyWhat2020}.

  For corpus-linguistics purposes specifically, the accuracy levels obtained here—0.98 on UPoS and 0.92 on LAS—are adequate for systematic querying of syntactic and morphological patterns, enabling analyzes on material that previously required laborious close reading.

  \subsection{Limitations and Constraints}
  \label{sec:limitations}

  There are several constraints that bear on interpretation and transferability. First, the sampling design was stratified only by sentence length and did not consult model uncertainty (Section~\ref{sec:sampling}). Second, a single annotator worked throughout, so no inter-annotator agreement exists and none can be recovered (Section~\ref{sec:consistency}). Third, no formal convergence criterion determined when to stop the loop and the analysis of where returns diminished is retrospective (Section~\ref{sec:stopping}). Fourth, each iteration involved both an increase in training data and a retraining from scratch, so the learning curve conflates data accumulation with restart effects (Section~\ref{sec:partitioning}). Finally, the word-embedding sets were rebuilt at two points where the curve exhibits steps (Section~\ref{sec:adp-dip}), introducing an additional source of variance.

  Three further limitations concern properties of the corpus, rather than the annotation pipeline, and they limit where the method is likely to be transferable.

  \paragraph{Dependence on formulaic structure.} The principal efficiency gains reported here arise from the formulaic nature of the documents (Section~\ref{sec:challenges}). Once the enumerative frame is learned, the model correctly predicts most items in a new batch before human correction. Corpora lacking such repetitiveness would not produce the same learning curve and correction rates would be expected to plateau earlier and at a higher residual error.

  \paragraph{Tractable code-switching.} The bilingual configuration in this corpus is asymmetric, with Latin providing the grammatical frame while vernacular forms appear chiefly as object names. That lexical concentration renders code-switching more tractable than cases where language alternation occurs at the clause level, or where the languages involved are typologically similar. The observed gains on code-switched sentences should therefore not be generalized to all multilingual historical materials.

  \paragraph{Effective synchronicity.} The target corpus spans under two centuries of a single documentary tradition from one city, and sampling ignored chronological stratification. This simplification is defensible for the present material, but not necessarily generalizable (Section~\ref{sec:future} discusses what a diachronically stratified version of the procedure would involve).

  \subsection{Future Research Directions}
  \label{sec:future}

  \subsubsection{Technical and Methodological Improvements}

  Five potential extensions to this work follow directly from what the study did not do.

  Active learning is the most obvious of them, and this case study is an unusually clean baseline for testing it. Because selection was held constant—\emph{i.e.} random within length strata, never consulting the model—the learning curve reported in this paper measures the effect of adding in-domain data as such, which makes it a lower bound rather than a demonstration of targeted intervention \autocite{holzingerInteractiveMachineLearning2016,amershiPowerPeopleRole2014,settlesActiveLearning2012,tomanekSemisupervisedActiveLearning2009}. Re-running the same nine rounds with uncertainty sampling, diversity sampling, or query-by-committee against the same gold standard would isolate the contribution of the selection policy exactly, since everything else could be held fixed \autocite{settlesActiveLearningLiterature2009}. A gain from targeted selection is plausible, however, it would need to be substantial to be practically significant, since at 0.50 minutes per sentence by $s_9$, halving the number of sentences saves less than one hour.

  The second area worth exploring is multi-task learning. In the present pipeline, Stanza treats tagging, lemmatization, and parsing as separate tasks. Residual tagging confusions (notably the noun–adjective boundary discussed in Section~\ref{sec:residual-errors}) could be disambiguated by syntactic context, suggesting that a joint training regime may be advantageous. A shared representation that exposes syntactic and lexical signals to one another could therefore reduce such errors and improve sample efficiency \autocite{caruanaMultitaskLearning1997,sogaardDeepMultitaskLearning2016}.

  A transformer-based encoder could also enhance the methodology. As noted in Section~\ref{sec:modern-baseline}, the experiment presented here used Latin models as currently distributed by Stanza, which does not ship a transformer-backed Latin package. Recent work supplies Latin-specific encoders that can be integrated into the same training pipeline, and whether a transformer backbone would materially reduce annotation demand or primarily raise the attainable performance ceiling is an important open question for follow-up work \autocite{manjavacasAdaptingVsPretraining2022}.

  Expanding the lexicon is the most direct remedy for the residual lemmatization errors identified in Section~\ref{sec:residual-errors}. Historical dictionaries, regional glossaries, and object-name inventories from museum and archaeological databases are natural sources, and resources such as LEMLAT demonstrate the value of dictionary-derived lexical material for Latin \autocite{passarottiLemlat30Package2017}. The principal challenge lies in harmonizing external lemmatization conventions with the UD schema used here, which would require careful mapping and substantial editorial effort.

  The annotation interface could also benefit from practical improvements. The general-purpose annotation environment used, \emph{brat}, lacks explicit support for orthographic variants and provides no persistent memory of analyzes applied to recurring formulae across batches. Incorporating features such as orthographic-normalization assistance, variant-aware lexicon lookups, and memoization of prior corrective decisions would reduce repetitive work and improve consistency on structurally repetitive corpora like this one (Section~\ref{sec:challenges}) \autocite{stenetorpBratWebbasedTool2012}.

  \subsubsection{Broader Application Domains}

  Beyond procedural refinements, three directions appear particularly promising for further study.

  The first is cross-linguistic replication, that is, running the same procedure on a documentary corpus in another language that already has an existing treebank. This is the most direct test of whether the quantitative outcomes observed here reflect properties of the method or idiosyncrasies of medieval Latin. The conditions listed in Section~\ref{sec:generalizability} convert the proposal into a concrete experimental design requiring only a comparable documentary corpus and an available out-of-domain treebank. Matching correction schedules and plateaux across languages would strengthen claims of method generalizability \autocite{piotrowskiNaturalLanguageProcessing2012}.

  The second is to extend the loop backwards to the image level. The pipeline currently begins from transcribed text, and producing reliable transcriptions represented a large portion of the project effort. The same correction-over-generation logic applies to handwritten-text recognition. An annotator correcting a transcription is simultaneously reading the manuscript and could plausibly correct both transcription and linguistic annotation in a single integrated pass. Investigating whether joint correction can improve both transcription accuracy and downstream parsing is an open and practical question for projects that must perform both tasks \autocite{kestemontScriptIdentificationMedieval2021}.

  The third direction is diachronic stratification. The corpus was treated as effectively synchronic by pooling material that spans under two centuries. Corpora with genuine temporal depth cannot be handled in that way because period-specific regularities become indistinguishable when pooled, and consequently a pooled model may learn neither. A stratified procedure that samples and trains period-specific models from a shared pool—or alternatively treats period as an explicit conditioning variable—would support the study of language change within specialized registers and technical vocabularies as well as potentially address questions of temporal transferability \autocite{yangOvercomingLanguageVariation2017,kulkarniStatisticallySignificantDetection2015,nevalainenHistoricalSociolinguisticsLanguage2017}.

  \section{Conclusion}

  This study has demonstrated that human-in-the-loop iterative training can produce usable NLP models for a documentary corpus that is not covered by existing resources. Relative to ITTB, the out-of-domain model that seeded the loop, tagging and morphological scores increase from 57–80\% to 92–98\%, while syntactic scores increase from 17–65\% to 82–95\%— moving performance from effectively unusable to research grade. Lemmatization is excluded from these ranges because, as noted in Section~\ref{sec:results}, it is chiefly supplied by the domain lexicon rather than learned by the model; attributing lexicon-driven gains to the iterative procedure would therefore be misleading.

  The result has a practical methodological implication for low-resource historical NLP. When no in-domain training data exist, and the target domain lies far from available corpora, constructing training data by correcting an imperfect model can be the most efficient route to a usable system. In this study, 33.2 hours of correction produced a parser that outperforms models trained on nearly two orders of magnitude more text. The critical requirement is domain match rather than sheer scale, and correction is a cost-effective means of achieving domain match compared with annotation from scratch \autocite{fortAmazonMechanicalTurk2011,wynneDevelopingLinguisticCorpora2005}.

  \subsection{Materials for Reuse}

  The corpus, the gold standard, the domain lexicon, the trained models for every iteration, and the accompanying code are released under open terms (Section~\ref{sec:availability}). Another project can therefore use the annotated data as a starting point, or re-run the evaluation against its own models, without reconstructing any of it from the description here.

  Two metrics in particular are intended as practical guidelines for comparable projects. First, the correction rate per round (Table~\ref{tab:annotation-effort-efficiency}) can be computed without a gold standard and tracked in real time. Based on the results presented in this paper, it aligns closely with evaluation metrics and therefore serves as a pragmatic stopping signal when constructing a gold standard is infeasible. Second, the annotation rate—reported together with the annotation scheme it covers—provides a replicable benchmark that other projects can measure themselves against for planning and cost estimation.

  \subsection{Data and Code Availability}
  \label{sec:availability}

  All materials necessary to reproduce the analyzes, tables, and figures presented in this paper are publicly available.

  The annotated corpus—1,804 sentences of manually corrected morphosyntactic annotation, together with the 200-sentence gold standard and the accompanying domain lexicon—is released in CoNLL-U format. The analysis code is released as a Python package which implements evaluation, lexicon construction, sampling, corpus preparation, and training-set construction. Every table and figure in this paper is generated from the released data by a single command, so these outputs can be regenerated and independently verified.\footnote{\url{https://github.com/gpizzorno/correction-as-annotation}}

  The trained models for all nine iterations are archived separately, since their size makes them unsuitable for a code repository.\footnote{\url{https://doi.org/10.5281/zenodo.22048602}}

  \printbibliography{}

\end{multicols}
\end{document}